\documentclass[sigconf, screen]{acmart}

\usepackage{subcaption}
\usepackage{dirtytalk}
\usepackage{lipsum}
\usepackage[ruled, linesnumbered]{algorithm2e}
\usepackage{xcolor, colortbl}
\usepackage{bbold}
\usepackage{multirow}
\usepackage{url}
\usepackage{arydshln}
\usepackage{todonotes}
\usepackage{hyperref}

\SetKwInput{KwInput}{Require}                
\SetKwInput{KwOutput}{Return}              
\SetKwComment{Comment}{/* }{ */}
\newcommand{\ourmethod}{\texttt{ForecastAD}}

\definecolor{myyellow}{rgb}{0.98, 0.98, 0.82}
\definecolor{mygreen}{rgb}{0.94, 1.0, 0.94}
\definecolor{tabBlue}{rgb}{0.90, 0.96, 0.97}
\definecolor{ColorNico}{rgb}{1.0, 0.66, 0.0}

\definecolor{deepblue}{rgb}{0.0, 0.0, 1.0}
\newcommand{\sukanya}[1]{{#1}}

\theoremstyle{plain}

\theoremstyle{definition}

\theoremstyle{remark}

\newtheorem*{assumption*}{\assumptionnumber}
\providecommand{\assumptionnumber}{}
\makeatletter

\AtBeginDocument{%
  \providecommand\BibTeX{{%
    \normalfont B\kern-0.5em{\scshape i\kern-0.25em b}\kern-0.8em\TeX}}}

\copyrightyear{2024} 
\acmYear{2024} 
\setcopyright{acmlicensed}\acmConference[KDD '24]{Proceedings of the 30th ACM SIGKDD Conference on Knowledge Discovery and Data Mining}{August 25--29, 2024}{Barcelona, Spain}
\acmBooktitle{Proceedings of the 30th ACM SIGKDD Conference on Knowledge Discovery and Data Mining (KDD '24), August 25--29, 2024, Barcelona, Spain}
\acmDOI{10.1145/3637528.3671623}
\acmISBN{979-8-4007-0490-1/24/08}

\begin{document}

\title{Detecting Abnormal Operations in Concentrated Solar Power Plants from Irregular Sequences of Thermal Images}

\author{Sukanya Patra}

\affiliation{%
  \institution{University of Mons}
  \city{Mons}
  \country{Belgium}
}
\email{sukanya.patra@umons.ac.be}

\author{Nicolas Sournac}
\affiliation{%
  \institution{University of Mons}
  \city{Mons}
  \country{Belgium}}
\email{nicolas.sournac@student.umons.ac.be}

\author{Souhaib Ben Taieb}
\affiliation{%
  \institution{University of Mons}
  \city{Mons}
  \country{Belgium}
}
\email{souhaib.bentaieb@umons.ac.be}








\begin{abstract}

\sukanya{Concentrated Solar Power (CSP) plants store energy by heating a storage medium with an array of mirrors that focus sunlight onto solar receivers atop a central tower. Operating at high temperatures these receivers face risks such as freezing, deformation, and corrosion, leading to operational failures, downtime, or costly equipment damage. We study the problem of anomaly detection (AD) in sequences of thermal images collected over a year from an operational CSP plant. These images are captured at irregular intervals ranging from one to five minutes throughout the day by infrared cameras mounted on solar receivers. Our goal is to develop a method to extract useful representations from high-dimensional thermal images for AD. It should be able to handle temporal features of the data, which include irregularity, temporal dependency between images and non-stationarity due to a strong daily seasonal pattern. The co-occurrence of low-temperature anomalies that resemble normal images from the start and the end of the operational cycle with high-temperature anomalies poses an additional challenge. We first evaluate state-of-the-art deep image-based AD methods, which have been shown to be effective in deriving meaningful image representations for the detection of anomalies. Then, we introduce a forecasting-based AD method that predicts future thermal images from past sequences and timestamps via a deep sequence model. This method effectively captures specific temporal data features and distinguishes between difficult-to-detect temperature-based anomalies. Our experiments demonstrate the effectiveness of our approach compared to multiple SOTA baselines across multiple evaluation metrics. We have also successfully deployed our solution on five months of unseen data, providing critical insights to our industry partner for the maintenance of the CSP plant. Our code\footnote{\href{https://github.com/sukanyapatra1997/ForecastAD}{https://github.com/sukanyapatra1997/ForecastAD}} is publicly accessible. Additionally, as our dataset is confidential, we release a simulated dataset\footnote{\href{https://tinyurl.com/kdd2024Dataset}{https://tinyurl.com/kdd2024Dataset}}.}
     
\end{abstract}





\maketitle

\section{Introduction}

The focus on renewable energies to counteract climate change has intensified recently. However, a critical challenge in adopting renewable energy sources is ensuring on-demand generation and dispatchability. A promising solution to this challenge is the integration of Thermal Energy Storage (TES) facilities, which temporarily store energy by heating or cooling a storage medium, such as water or molten salt. Concentrated Solar Power (CSP) plants effectively utilize TES for storing energy by heating the medium with an array of mirrors focused on solar receivers atop a central tower \cite{Zhang2013ConcentratedMethodology}. These solar receivers are composed of vertical heat exchanger tubes arranged in panel form, allowing the medium to flow through them.

Operating at extreme temperatures, these systems are prone to adverse effects, including the freezing of the medium (affecting a subset of vertical tubes with significantly higher temperatures), damage to heat-resistant coatings, and deformation and corrosion of the heat exchanger tubes. Therefore, meticulous monitoring of the process is crucial. Given the vast amount of data generated from multiple sensors, manually detecting abnormal behaviours becomes impractical. This necessitates an automated system capable of immediately identifying abnormal behaviours. The advantages of such a system are twofold: it ensures smooth operation and uninterrupted power generation by minimizing downtime, and it reduces the risk of further equipment damage by allowing for prompt failure responses. This approach also leads to an extended operational lifetime for the CSP plant.

In this paper, \sukanya{our goal is to develop a deep image-based anomaly detection (AD) \cite{Ruff2021ADetection, Pang2020DeepReview} method to identify abnormal behaviours in sequences of thermal images collected over a span of one year from an operational CSP plant.} These images are captured at irregular intervals ranging from one to five minutes throughout the day by infrared cameras mounted on solar receivers. Our problem is related to data-driven Predictive Maintenance (PdM), where the state of equipment in industrial processes is monitored to predict future failures \cite{Tang2020DeepImages}.

Specifically, we aim to develop an AD method capable of extracting useful representations from high-dimensional thermal images. It should be able to handle temporal features of the data, which include irregularity, temporal dependency between images and non-stationarity due to a strong daily seasonal pattern. An additional challenge is the coexistence of low-temperature anomalies that resemble low-temperature normal images from the start and the end of the operational cycle alongside high-temperature anomalies. This necessitates learning the current state of the operational cycle to correctly identify anomalous operations.

We first examine the performance of state-of-the-art (SOTA) deep AD methods that have been successful in extracting useful image representations for anomaly detection, such as CFlow \cite{Gudovskiy2021CFLOW-AD:Flows}, PatchCore \cite{Roth2022TowardsDetection}, and DR{\AE}M \cite{Zavrtanik2021DRMDetection}. Our experiments confirm that neglecting the temporal features of the data leads to low accuracy, especially in distinguishing low-temperature normal samples from anomalies. Then, we explore a new forecasting-based AD method, \ourmethod{}, which predicts the image for a given future time based on a sequence of past observed images and their timestamps using a deep sequence model. \ourmethod{} extracts relevant representations from the high-dimensional images and captures the normal behaviour of the solar receivers, taking into account the temporal features of the data. An anomaly is then defined as a significant deviation from the learned normal behaviour. Our experiments demonstrate the effectiveness of \ourmethod{} compared to multiple SOTA baselines across various evaluation metrics. We have also successfully deployed our solution on five months of unseen data, providing critical insights for the maintenance of the CSP plant.

\sukanya{We first discuss related work in Section~\ref{sec:related_work}. Secondly, in Section~\ref{sec:usecase}, we present the case study on detecting anomalous behaviours in CSP plants. Then, we explain our forecasting-based approach, \ourmethod{} in Section~\ref{sec:forecast}. Finally, in Section~\ref{sec:eval}, we discuss our empirical results before providing our concluding remarks in Section~\ref{sec:conclusion}.}

\section{Related Work}
\label{sec:related_work}
AD has been extensively studied over several decades \cite{Ruff2021ADetection}. The major AD methods can be broadly classified into four categories - density-based, reconstruction-based, classification-based approaches, and feature embedding-based methods.\vspace{-0.2cm}\\

\noindent \textbf{Density-based methods}. The density-based approach aims to estimate the probability distribution of normal data by assuming that normal samples are more likely to occur under the estimated distribution than anomalous samples. Traditional methods \cite{Kind2009Histogram-basedDetection, Parzen1962OnJSTOR, Bishop1994NoveltyValidation} fit a model to arbitrary data distribution but encounter challenges in higher-dimensional input spaces due to the curse of dimensionality. To overcome this, they are often applied to low-dimensional latent representations obtained using techniques like Autoencoder (AE) and Variational Autoencoder (VAE). Neural generative models, such as VAE and Generative Adversarial Networks (GANs), are deep learning-based methods that estimate a neural network's parameters to map a predetermined source distribution to the input data distribution. Recent AD methods such as CFlow \cite{Gudovskiy2021CFLOW-AD:Flows} and FastFlow \cite{Yu2021FastFlow:Flows} further build on normalizing flows. However, studies demonstrated that normalizing flows often struggle to detect anomalies and assign them a higher likelihood \cite{Nalisnick2018DoKnow, Nalisnick2019DetectingTypicality, Kirichenko2020WhyData}.\vspace{-0.2cm}\\

\noindent \textbf{Reconstruction-based methods}. Reconstruction-based methods operate on the assumption that encoder-decoder models trained on normal samples will exhibit poor performance for anomalous samples. Common deep reconstruction models used include AE or VAE-based approaches, while advanced strategies involve reconstruction by memorized normality \cite{Gong2019MemorizingDetection}, model architecture adaptation \cite{Lai2019RobustDetection} and partial/conditional reconstruction \cite{Yan2021LearningDetection, Nguyen2019AnomalyPredictions}. Recent approach DR{\AE}M \cite{Zavrtanik2021DRMDetection} trains a discriminative network alongside the reconstruction network to localize anomalies without the need for any further post-processing steps. Generative models like GANs are also widely employed for anomaly detection, as the discriminator inherently calculates reconstruction loss for samples \cite{Zenati2018EfficientDetection}.\vspace{-0.2cm}\\

\noindent \textbf{One-class classification}. Anomaly detection can be approached as a one-class classification \cite{Tax2001One-ClassCounter-Examples, Khan2014One-classTechniques} or single-class classification \cite{Minter1975Single-ClassClassification, El-Yaniv2006OptimalStrategies} problem. Unlike density-based methods, classification-based techniques, such as One-Class Support Vector Machine (OC-SVM) \cite{Scholkopf2001EstimatingDistribution}, directly estimate a decision boundary to differentiate between normal and anomalous samples. However, this task can be challenging due to imbalanced datasets, where normal samples vastly outnumber anomalous samples. To address this, techniques like Support Vector Data Descriptor (SVDD) \cite{Tax1999SupportDescription, Tax2001One-ClassCounter-Examples, Tax2004SupportDescription} derive a tight spherical bound. To enhance the expressivity of the classical models, deep learning models are used to learn the features from the data \cite{Ruff2018DeepClassification,Erfani2016High-dimensionalLearning}.\vspace{-0.2cm}\\

\noindent \textbf{Feature Embedding-based methods} There are mainly two different types of feature embedding-based anomaly detection methods: memory bank \cite{Defard2021PaDiM:Localization, Roth2022TowardsDetection, Lee2022CFA:Localization}, student-teacher \cite{Zhang2023ContextualDetection, Batzner2024EfficientAD:Latencies}. The main idea of memory bank methods is to extract features of nominal images and store them in a memory bank during the training phase. During the testing phase, the feature of a test image is used as a query to match the stored nominal features. The performance of the memory bank methods heavily depends on the completeness of the memory bank requiring a large number of nominal images. In the student-teacher approaches \cite{deng2022anomaly, Zhang2023ContextualDetection}, the student network learns to extract features of the nominal samples, similar to the teacher model. For anomalous images, the features extracted by the student network should be different from the teacher network.

\vspace{-0.5cm}
\section{A Case Study on Detecting Anomalous Behaviours in CSP Plants}
\label{sec:usecase}

\begin{figure*}[t]
\begin{subfigure}[b]{0.45\textwidth}
 \centering
  \includegraphics[width=\textwidth]{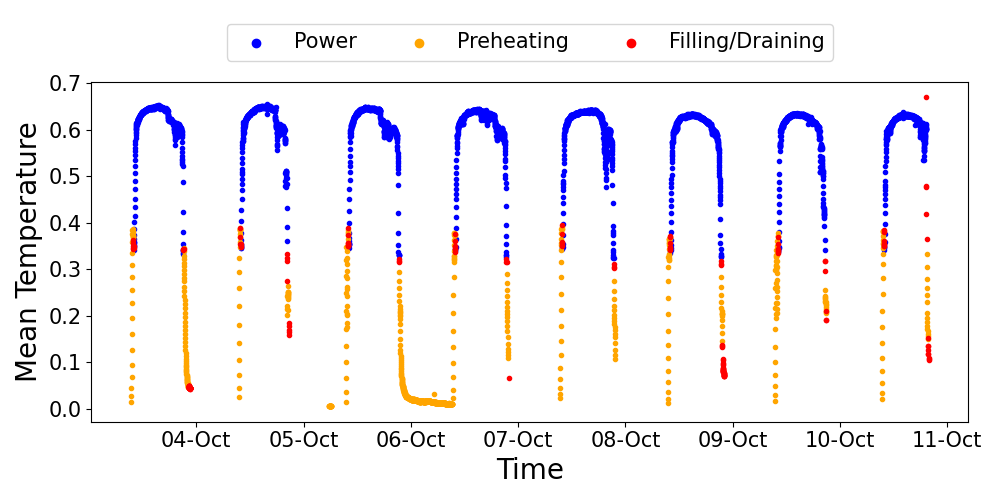}\vspace{-0.2em}
    \caption{Average image temperature for multiple days.}
    \label{fig:mean}
\end{subfigure}
\begin{subfigure}[b]{.49\textwidth}
 \centering
  \includegraphics[width=0.8\textwidth]{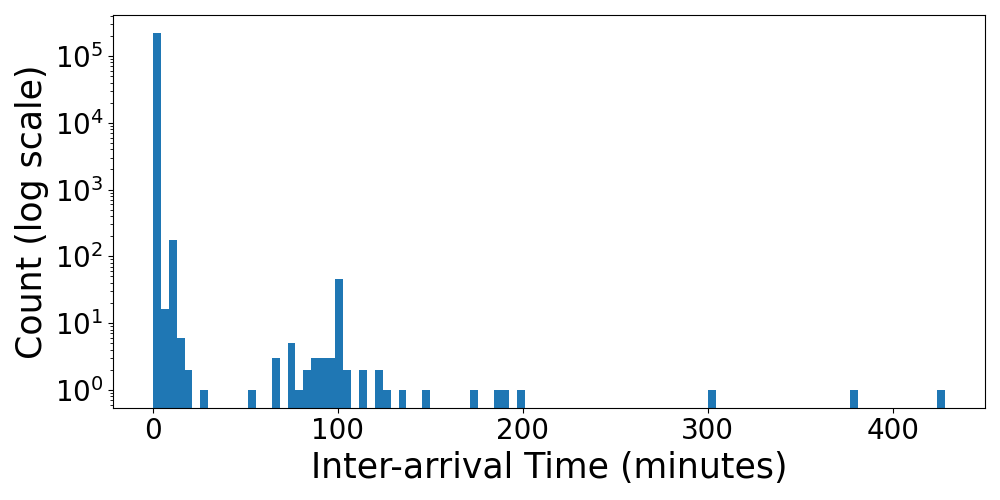}\vspace{-0.2em}
    \caption{Histogram of inter-arrival times}
    \label{fig:inter_hist_nor}
\end{subfigure}\vspace{-1em}
\caption{Visualisation of different properties of the data}
\label{fig:data_properties}
\end{figure*}

A CSP plant consists of two main components, namely: (i) the Thermal Solar Receiver and (ii) the Steam Generator. The Thermal Solar Receiver placed on top of a central tower in the plant acts as a solar furnace. On the ground surrounding the tower, an array of flat, movable mirrors called heliostats concentrate the sun rays on the solar receiver. The receiver consists of vertical heat exchanger tubes through which the heat transfer medium flows, absorbing the heat from the concentrated sun rays. Then, the absorbed thermal energy is utilized to generate superheated steam, which runs the Steam Generator for the production of energy. In this work, we focus on detecting anomalous behaviours of the Thermal Solar Receiver using data obtained from an operational plant.

CSP plants utilize high-capacity fluids like molten salts as the heat transfer medium, which are stored in TES facilities for future use. This allows for the on-demand generation of energy, making CSP plants a viable alternative to fossil fuel-based energy plants. However, due to operation in extreme temperatures, the solar thermal receivers are adversely impacted in several ways:\vspace{-0.2cm}\\

\noindent\textbf{[i] Blocked Tubes}. The molten salts passing through the heat exchanger tubes tend to freeze in localized zones when the temperature falls below a certain threshold, blocking them.\vspace{-0.2cm}\\

\noindent\textbf{[ii] Deformity}. The metal heat exchanger tubes in the receiver tend to expand due to the high temperatures. Uneven dilation of the tubes could eventually lead to deformity.\vspace{-0.2cm}\\

\noindent\textbf{[iii] Stress and Metal Fatigue}. The metal tubes in the receiver undergo expansion when exposed to high temperatures during regular operation and contraction when the operation ends. Such repeated changes lead to metal fatigue. Additionally, the pressure generated from the flowing molten salts exerts stress on the tubes.\vspace{-0.2cm}\\

\noindent\textbf{[iv] Corrosion}. Due to the interaction of the metal with the molten salt flowing through the tubes, it tends to deteriorate over time. These reactions are further accelerated due to the high temperatures in the receiver.\vspace{-0.2cm}\\

Hence, CSP plants require close monitoring to guarantee seamless operation and continuous power generation. Achieving this requires the analysis of data collected by numerous sensors installed on the Solar Receiver. Yet, the vast volume of data generated renders manual inspection unfeasible, thus underscoring the need for an automated, data-driven monitoring system.

\begin{figure*}[t]
\begin{subfigure}[b]{\textwidth}
 \centering
  \includegraphics[width=\textwidth]{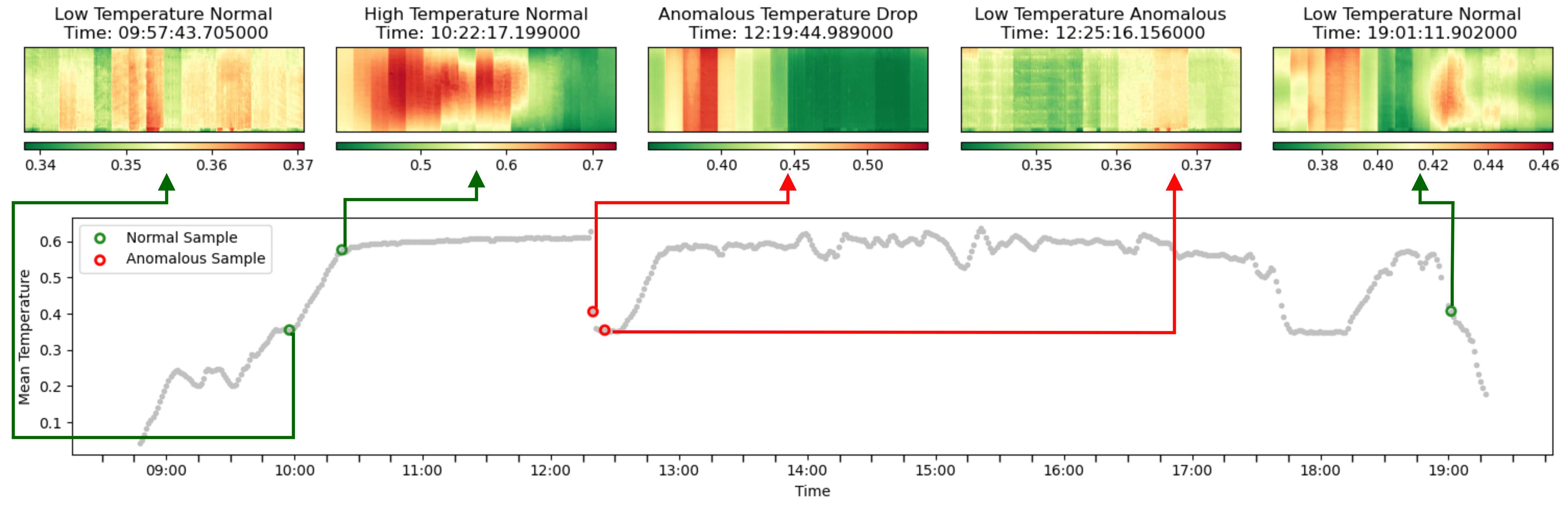}\vspace{-1em}
  \caption{Different types of normal and anomalous examples in a day}
  \label{fig:mean_samples}
\end{subfigure}
\centering
\begin{subfigure}[b]{\textwidth}
 \centering
  \includegraphics[width=\textwidth]{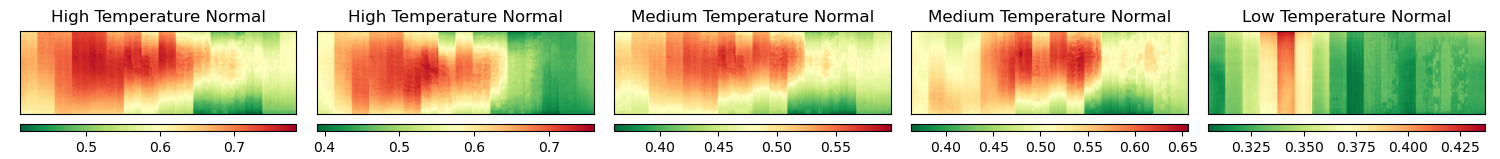}
  \caption{Normal examples}\vspace{-1em}
  \label{fig:normal}
\end{subfigure}
\begin{subfigure}[b]{\textwidth}
 \centering
  \includegraphics[width=\textwidth]{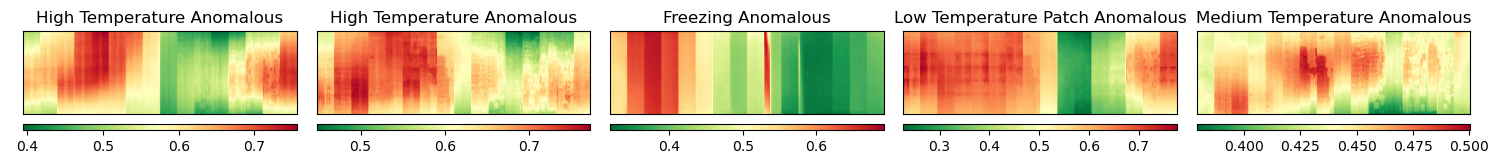}
  \caption{Anomalous examples}\vspace{-1em}
  \label{fig:anomalous}
\end{subfigure}
\caption{Examples of different types of normal and anomalous images}
\label{fig:examples}
\end{figure*} 

\subsection{Data Description}
\label{sec:datadscr}

The Thermal Solar Receiver is composed of several panels, each featuring vertical heat exchanger tubes. These tubes allow the heat transfer medium to flow through, effectively absorbing heat from the concentrated sunlight. Infrared (IR) cameras, strategically positioned around the solar receiver, capture the surface temperature, producing thermal images with dimensions of $184\times 608$. These images are captured approximately every one to five minutes, with each image's timestamp recorded. During normal operations, the temperature of the heat transfer medium gradually increases as it traverses the vertical tubes from one end of the panel to the other, a direct result of absorbing heat from the concentrated sunlight. Consequently, the surface temperature patterns recorded by the IR cameras are anticipated to exhibit a smooth gradient, aligning with the medium's flow direction. Our dataset covers a year of operational data without ground truth labels for the images (normal or abnormal), making it an unsupervised anomaly detection problem. Note that throughout this work, we provide normalized images from the dataset for the sake of confidentiality.

Operation in CSP plants occurs across three distinct phases as depicted in Figure~\ref{fig:mean}: (i) \emph{Preheating}, (ii) \emph{Filling/Draining}, and (iii) \emph{Power}. The molten salt used in CSP plants freezes when the temperature drops below a certain threshold. To avoid this, solar panels are initially heated during the \emph{Preheating} phase. Then, in the subsequent \emph{Filling/Draining} phase, molten salt is circulated within the panels. The \emph{Power} phase initiates as the molten salt absorbs heat from sunlight, facilitating power generation. As operations conclude, the molten salt is drained from the panels during the \emph{Filling/Draining} phase. Consequently, the panels commence cooling down, transitioning back to the \emph{Preheating}. \sukanya{Our work focuses solely on the \emph{Power} phase, as it is crucial for power generation and susceptible to damage from prolonged exposure to high temperatures.}\\

\noindent \textbf{Data characteristics and modelling challenges}. Through extensive data analysis, we identified the following additional challenges, which are essential for modelling the solar receiver data:\vspace{0.55mm}\\

\noindent\textbf{[i] Non-stationarity}. Figure \ref{fig:mean} presents the average surface temperature across a specific week, highlighting temporal variations in the mean image temperature and demonstrating a clear pattern of daily seasonality in the data.\vspace{-0.2cm}\\

\noindent\textbf{[ii] Irregular sampling.} The images were captured at irregular time intervals, as illustrated in Figure \ref{fig:inter_hist_nor}, which depicts the distribution of inter-arrival times. Additionally, the dataset lacks data for the extended periods when the plant was not operational.\vspace{-0.2cm}\\

\noindent\textbf{[iii] Temporal dependence.} The images exhibit a strong temporal dependence, influenced significantly by weather conditions.\vspace{-0.2cm}\\

\noindent\textbf{[iv] High dimensionality}. \sukanya{Anomalous characteristics often stay hidden and unnoticed due to data sparsity in high-dimensional spaces. Identifying features that capture the essential high-order, non-linear interactions needed for AD is thus challenging.}\vspace{-0.2cm}\\ 

\noindent\textbf{[v] Large volume of data}. The dataset comprises images captured throughout a year of operation at approximately one to five-minute intervals, leading to a vast volume of data.\vspace{-0.2cm}\\

\noindent\textbf{[vi] Unlabeled data}. Our dataset lacks ground truth labels for the images, whether they are normal or abnormal, classifying our task as an unsupervised anomaly detection problem.

\subsection{Data Labelling}

\begin{figure*}[t]
    \centering
    \includegraphics[width=0.85\textwidth]{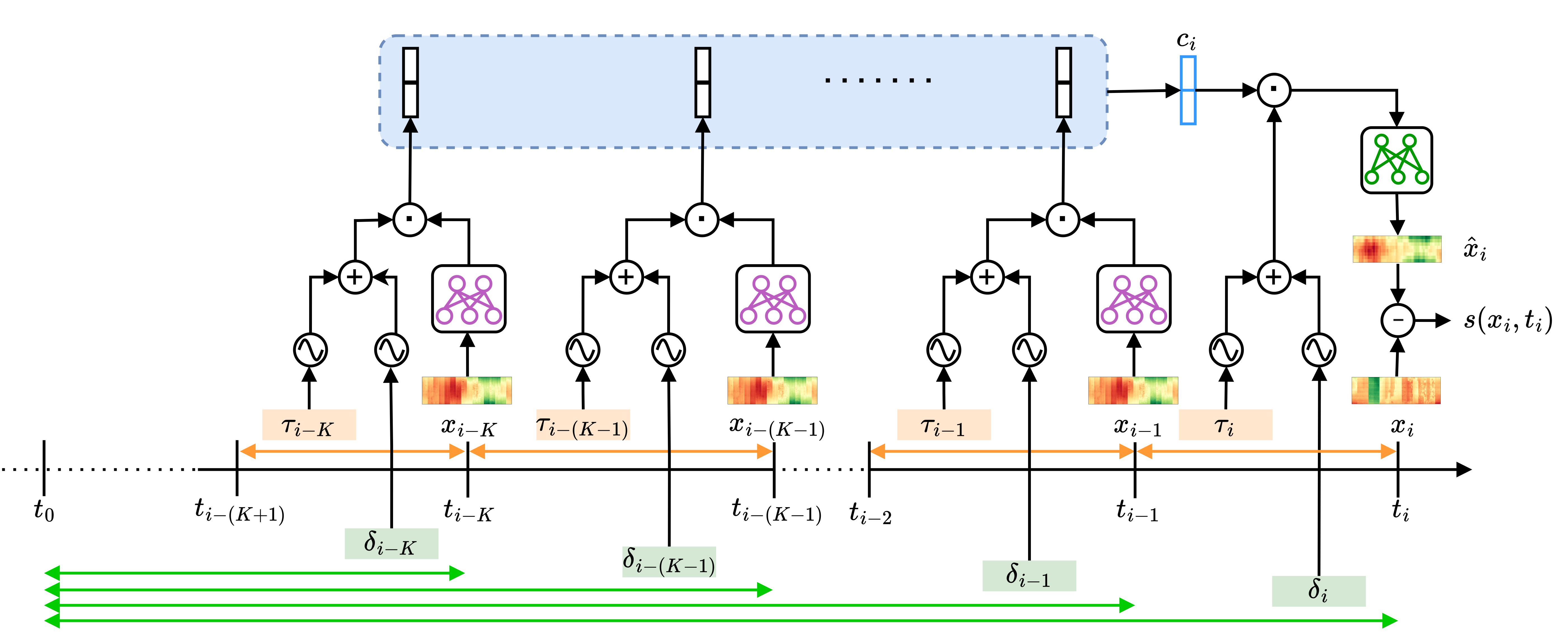}
    \caption{Illustration of the end-to-end architecture of \ourmethod{}. The model is trained to forecast the next image in the sequence given a context embedding $c_i$ of $K$ prior data points obtained using a sequence-to-sequence model. For $(x_{i-k}, t_{i-k}, y_{i-k}) \in \mathcal{D}$ in the context, we sum the embeddings of inter-arrival time $\tau_{i-k}$ and interval since the start of operation $\delta_{i-k}$ and concatenate it with the image embedding. The anomaly score \(s(x_{i}, t_{i})\) is computed as the difference between the forecasted and original image.}
    \label{fig:forecast}
\end{figure*}

To effectively assess the performance of various AD methods, we have labelled a subset of data from the CSP plant. This endeavour is notably complex due to the plant's operation across multiple phases, each characterized by unique temperature ranges. Consequently, this diversity leads to a range of normal and anomalous sample types, as depicted in Figure~\ref{fig:examples}. The challenge of identifying anomalies through the plant's operational phases is evident in Figure~\ref{fig:mean_samples}. Notably, normal images with low temperatures at the operation's start and end (the left-most and right-most images in Figure~\ref{fig:mean_samples}) closely resemble low-temperature anomalies (the second image from the right in Figure~\ref{fig:mean_samples}). The distinction between these samples relies heavily on context.

Moreover, the variable nature of anomalies adds a layer of complexity to the labelling process. Our approach to this challenge is informed by a deep understanding of the CSP plant's operations and expert insights from the field. We categorize the \emph{Power} phase into three distinct segments: (i) Starting (S), where the solar receiver's mean temperature begins to rise; (ii) Middle (M), where it reaches and maintains its peak; and (iii) Ending (E), where it declines as the day concludes. In our preprocessing, we exclude days with significantly few samples or with a consistently low temperature throughout the M segment, likely indicative of sensor or system failures. For the S and E segments, samples showing a consistent temperature increase (> $5^{\circ}$C) or decrease (< -$5^{\circ}$C), respectively, are deemed normal, whereas those displaying contrary trends are marked as anomalous. In the M segment, we apply the following four rules for labelling:\\

\noindent[\textbf{R1}.] \textbf{Difference between consecutive images}. \sukanya{During the M segment of the \emph{Power} phase, we expect a stable temperature. Significant deviations from the preceding observation indicate an abnormality. To detect such anomalies, we first compute the pixel-wise squared differences between every two consecutive images. For each pair, we select the 95th percentile of these pixel-wise differences as our \textit{score}. Samples are then labelled as anomalous if their score exceeds the 99.9th percentile of the scores for all samples in the dataset}\vspace{-0.2cm}\\

\noindent[\textbf{R2}.] \textbf{Difference from average daily temperature}. \sukanya{Samples with average temperatures that significantly deviate from the daily average temperature are labelled as anomalous. To identify these anomalies, we first compute the daily mean temperature. Then, we calculate the difference between each image's average temperature and the mean temperature of the corresponding day, which serves as our \textit{score}. Finally, samples are labelled as anomalous if their score falls below the 1st percentile of the distribution of scores across all samples in the dataset.}\vspace{-0.2cm}\\

\noindent[\textbf{R3}.] \textbf{Difference with specific daily normal samples.} Rules \textbf{R1} and \textbf{R2} are limited to the detection of low-temperature anomalous samples. \sukanya{To address this, we select the first five images from the M segment of \emph{Power} phase of each day to serve as a set of templates for that day. We then employ a similar methodology as in Rule \textbf{R1}, but instead of comparing an image to just the prior image, we compute the mean difference between the image and all five templates of the corresponding day.} Applying this rule allows us to obtain sets of high-temperature normal and abnormal samples, along with a diverse set of low-temperature abnormal samples.\vspace{-0.2cm}\\

\noindent[\textbf{R4}.] \textbf{Freezing statistics.} To identify the anomalous samples with characteristics such as freezing and low-temperature patches, \sukanya{we compute row-wise and column-wise differences within each image. First, we calculate the maximum value of the element-wise differences between two consecutive rows, which we term the \textit{horizontal score}. Next, we compute the element-wise differences between consecutive columns and apply a Sobel filter \cite{sobel1968isotropic} to detect vertical edges. The mean value of the elements detected by the Sobel filter across all columns is referred to as the \textit{vertical score}. An image is labelled as anomalous if either the horizontal or the vertical score exceeds a predefined threshold.}

\sukanya{Given the labelling rules, we first apply them to obtain an initial set of labels. Then, in collaboration with domain experts, we conduct a visual analysis of the labelled thermal images. For the visual analysis, we also perform clustering on the labelled samples and inspect the cluster centres in addition to analyzing each image individually. This thorough inspection leads to subsequent refinements of the labelled set, enhancing the reliability of the labels for accurately assessing anomaly detection methods.}

\begin{figure*}[h]
    \centering
    \includegraphics[width=0.9\textwidth]{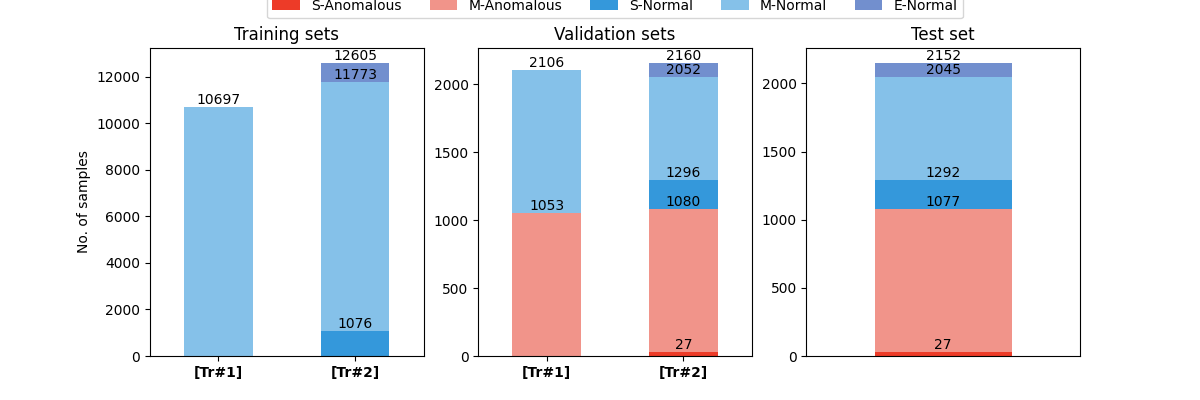}
    \vspace{-1em}
    \caption{Dataset split for two different training setups and the test set}
    \label{fig:training_setup}
\end{figure*}

\subsection{General problem formulation}

\sukanya{Consider a dataset \( \mathcal{D} = \{(x_i, t_i, y_i)\}_{i=1}^n \) consisting of \( n=16,917 \) triplets.} Each \( x_i \in \mathcal{X} = \mathbb{R}_{+}^d \) corresponds to a thermal image with dimension \( d=H \times W \), where the height \( H \) is \( 184 \) and the width \( W \) is \( 608 \). These images were captured at times \( t_i \in \mathbb{R}_{+} \), and each \( y_i \in \{0,1\} \) denotes the corresponding label, with \( 0 \) representing the normal class and \( 1 \) representing the anomalous class.

Let \( \mathcal{D}_N \), \( \mathcal{D}_V \) and \( \mathcal{D}_T \) denote disjoint training, validation and test sets, respectively, with \( \mathcal{D}_N \cup \mathcal{D}_V \cup \mathcal{D}_T = \mathcal{D} \). \( \mathcal{D}_N \) is exclusively composed of normal samples, i.e., \( y_i = 0 \) for all \( (x_i, t_i, y_i) \in \mathcal{D}_N \). \( \mathcal{D}_V \) and \( \mathcal{D}_T \) include both normal and anomalous samples.

Using the training set \( \mathcal{D}_N \), \sukanya{the AD methods aim to learn} a \sukanya{scoring function \( s(\cdot, \cdot): \mathbb{R}_{+}^d \times \mathbb{R}_{+} \rightarrow \mathbb{R} \) that assigns an anomaly score \( s(x, t) \) to any given point \( (x, t) \).} By using a threshold \( \lambda \in \mathbb{R} \), this anomaly score can then be converted into a predicted label \( \hat{y} \) as follows:
\begin{equation}
    \hat{y} =  
\begin{cases}
    1, & \text{if } s(x, t) \geq \lambda;\\
    0, & \text{if } s(x, t) < \lambda.
\end{cases} 
\end{equation}

\section{A forecasting-based AD model}
\label{sec:forecast}

We present a new forecasting-based AD method, denoted \ourmethod{}, to detect anomalous operations in the Thermal Solar Receiver of a CSP plant from irregular sequences of thermal images. The proposed method builds a forecasting model to reconstruct the thermal images using past observations as context. Images that are hard to reconstruct are considered anomalous. For a given image, our procedure can be summarized in the following steps: (i) extract feature embeddings for that image (Section~\ref{sec:fe}), (ii) use the previous $K$ images as \emph{context} and encode them using a deep sequence model (Section~\ref{sec:hist}), and (iii) using the context, reconstruct the image with a decoder forecasting model (Section~\ref{sec:dec}), then assign an anomaly score based on the reconstruction error between the original and predicted image. We provide an overview of the architecture of \ourmethod{} in Figure~\ref{fig:forecast} and summarize it in Algorithm~\ref{algo:forecastAD}.

\subsection{Image Encoder}
\label{sec:fe}

Using the training data, $\mathcal{D}_N$, we pre-train an encoder network to capture the inherent structure of our dataset's images. The image encoder, denoted by \( \phi_e(\cdot; W_e): \mathcal{X} \rightarrow \mathcal{Z} \), transforms images from the high-dimensional input space \( \mathcal{X} \) to a compact latent space \( \mathcal{Z} = \mathbb{R}^{d'} \), significantly reducing dimensionality where \( d' << d \). We use an autoencoder framework for image reconstruction, with a decoder network \( \phi_d(\cdot; W_d): \mathcal{Z} \rightarrow \mathcal{X} \) to project images from the latent space \( \mathcal{Z} \) back to the original input space \( \mathcal{X} \). The autoencoder is given by \( \phi = \phi_e \circ \phi_d \), with \( \circ \) indicating function composition. \sukanya{Given the high dimensionality of the input image}, we opt for a multi-layer deep convolutional network as the image encoder, exploiting its effectiveness in extracting meaningful representations directly from the data \cite{bishop2006pattern}. We calculate the reconstruction loss between original data points \( x_i \) and \sukanya{their} reconstructions \( \Tilde{x}_i = \phi(x_i) \) as:
\begin{equation}
    \mathcal{L}_{\mathrm{pre-train}} = \frac{1}{|\mathcal{D}_N|} \sum_{i=1}^{|\mathcal{D}_N|} \|x_i - \Tilde{x}_i\|_F^2,
\end{equation}
where \( \|\cdot\|_F \) denotes the Frobenius norm.

\subsection{Context Encoder}
\label{sec:hist}

To \sukanya{handle the irregular inter-arrival times \(\tau_i = t_i - t_{i-1}\) between successive \((i)\)-th and \((i-1)\)-th images}, our deep sequence model incorporates both the image sequences and their associated irregular inter-arrival times. We employ a sinusoidal encoding \(\psi_i = f_{\mathrm{sin}}(\tau_i)\), inspired by the positional encoding technique in transformer models \cite{Vaswani2017AttentionNeed}. This method aligns with strategies used in Neural Temporal Point Processes \cite{Enguehard2020NeuralRecords}. 

In addition to the inter-arrival times between consecutive images, we also embed the relative time since the start of the operation \(t_0\), i.e., \(\delta_i = t_i - t_0\), \sukanya{which provides information about the position of an image \(x_i\) within the operational cycle. Such temporal context helps in detecting challenging temperature-based anomalies, as it helps distinguish between low-temperature anomalies occurring mid-cycle and low-temperature normal images at the start of the operation.} We use the same sinusoidal encoding for the interval \(\delta_i\) as \(\Psi_i = f_{\mathrm{sin}}(\delta_i)\). The sum of the two time embeddings \(\psi_i\) and \(\Psi_i\) is combined with the image embedding to obtain the final embedding \(\hat{z}_i = [z_i \oplus (\psi_i + \Psi_i) ]\), where \(z_i = \phi_e(x_i)\) represents the image embedding and \(\oplus\) denotes the concatenation operator.

We compactly encode the embeddings of the \(K\) samples preceding the image at timestep \(t_{i}\) into a fixed-dimensional vector \(c_{i}\), termed the context vector for the \(i\)-th image. \sukanya{This can be accomplished with a deep sequence model.  In our implementation, we opt for an LSTM \(\varphi(\cdot; W_c)\), parameterized by \(W_c\).  For a given context sequence \(\mathcal{C}_{i} = \{\hat{z}_{i-K}, \cdots, \hat{z}_{i-1}\}\), the hidden state is recursively updated from previous states as \(c_{i} = \varphi(c_{i-1}, \hat{z}_{i-1}; W_c)\), starting from a random state.} 

\begin{table*}[t]
    \centering
    \caption{Anomaly detection performance. Style: best in bold and second best using underline}\vspace{-1em}
    \label{tab:res}
    \aboverulesep = 0pt
    \belowrulesep = 0pt
    \renewcommand{\arraystretch}{1.2}
    \resizebox{0.8\linewidth}{!}{
    \begin{tabular}{@{}c|l|ccc|ccc@{}}
    \toprule
    \multirow{2}{*}{\textbf{Train Setting}} & \multirow{2}{*}{\textbf{Model}} &  \multicolumn{3}{c|}{\textbf{AUROC} (\%)} &  \multicolumn{3}{c}{\textbf{AUPR} (\%)}\\
    &  & \textbf{[Ts\#1]} & \textbf{[Ts\#2]} & \textbf{[Ts\#3]} & \textbf{[Ts\#1]} & \textbf{[Ts\#2]} & \textbf{[Ts\#3]}\\
    \midrule 
    \multirow{8}{*}{\textbf{[Tr\#1]}} 
    &Autoencoder & 98.05 ($\pm$ 0.74) & 46.43 ($\pm$ 1.61) & 87.87 ($\pm$ 0.26) & 98.50 ($\pm$ 0.54) & 6.62 ($\pm$ 0.19) & 81.46 ($\pm$ 0.55) \\
    &CFlow \cite{Gudovskiy2021CFLOW-AD:Flows} & 94.68 ($\pm$ 1.26) & 39.99 ($\pm$ 2.33) & 82.91 ($\pm$ 1.08) & 96.28 ($\pm$ 0.92) & 5.94 ($\pm$ 0.24) & 76.11 ($\pm$ 1.04) \\
    &DR{\AE}M \cite{Zavrtanik2021DRMDetection}& 97.70 ($\pm$ 0.77) & 40.48 ($\pm$ 2.15) & 87.38 ($\pm$ 0.61) & 97.97 ($\pm$ 0.92) & 6.13 ($\pm$ 0.27) & 82.05 ($\pm$ 0.57) \\
    & FastFlow \cite{Yu2021FastFlow:Flows} & 99.83 ($\pm$ 0.03) & 47.32 ($\pm$ 0.29) & \textbf{91.36 ($\pm$ 0.25)} & 99.87 ($\pm$ 0.02) & \textbf{9.42 ($\pm$ 1.18)} & \underline{86.02 ($\pm$ 0.52)} \\
    &PaDiM \cite{Defard2021PaDiM:Localization} & \underline{99.85 ($\pm$ 0.02)} & \underline{49.86 ($\pm$ 0.47)} & \underline{91.23 ($\pm$ 0.10)} & \textbf{99.89 ($\pm$ 0.01)} & \underline{7.73 ($\pm$ 0.18)} & \textbf{87.25 ($\pm$ 0.21)} \\
    &PatchCore \cite{Roth2022TowardsDetection} & 99.23 ($\pm$ 0.08) & \textbf{50.58 ($\pm$ 0.37)} & 89.04 ($\pm$ 0.30) & 99.43 ($\pm$ 0.05) & 7.25 ($\pm$ 0.08) & 83.41 ($\pm$ 0.27) \\
    &Reverse Distillation \cite{deng2022anomaly} & 93.88 ($\pm$ 1.13) & 41.31 ($\pm$ 2.19) & 84.61 ($\pm$ 1.54) & 95.47 ($\pm$ 0.79) & 6.08 ($\pm$ 0.30) & 80.70 ($\pm$ 1.37) \\
    &\cellcolor{tabBlue}\ourmethod{} & \cellcolor{tabBlue}\textbf{99.86 ($\pm$ 0.05)} & \cellcolor{tabBlue}46.22 ($\pm$ 1.06) & \cellcolor{tabBlue}89.89 ($\pm$ 0.35) & \cellcolor{tabBlue}\textbf{99.89 ($\pm$ 0.04)} & \cellcolor{tabBlue}6.57 ($\pm$ 0.09) & \cellcolor{tabBlue}85.75 ($\pm$ 0.65) \\
    \midrule
    \multirow{8}{*}{\textbf{[Tr\#2]}} 
     &Autoencoder &  \underline{96.67 ($\pm$ 0.77)} & 45.92 ($\pm$ 2.47) & 85.45 ($\pm$ 1.18) & 96.91 ($\pm$ 0.93) &  6.69 ($\pm$ 0.35) & 78.61 ($\pm$ 1.30) \\
     &CFlow \cite{Gudovskiy2021CFLOW-AD:Flows} &  84.91 ($\pm$ 2.72) & 42.90 ($\pm$ 2.71) & 77.38 ($\pm$ 2.98) & 88.18 ($\pm$ 2.02) &  6.51 ($\pm$ 0.39) & 74.80 ($\pm$ 3.24) \\
     &DR{\AE}M \cite{Zavrtanik2021DRMDetection} &  93.52 ($\pm$ 0.52) & 40.51 ($\pm$ 1.33) & 85.71 ($\pm$ 0.78) & 94.56 ($\pm$ 0.44) &  7.62 ($\pm$ 1.01) & 83.36 ($\pm$ 1.08) \\
     &FastFlow \cite{Yu2021FastFlow:Flows} &  92.38 ($\pm$ 0.72) & 52.51 ($\pm$ 1.09) & 89.92 ($\pm$ 0.68) & 93.46 ($\pm$ 0.60) &  8.87 ($\pm$ 0.46) & 88.76 ($\pm$ 0.56) \\
     &PaDiM \cite{Defard2021PaDiM:Localization} &  95.99 ($\pm$ 0.37) & 58.14 ($\pm$ 1.00) & \underline{92.28 ($\pm$ 0.32)} & 96.77 ($\pm$ 0.32) & \underline{11.50 ($\pm$ 0.86)} & \underline{90.73 ($\pm$ 0.42)} \\
     &PatchCore \cite{Roth2022TowardsDetection} &  \textbf{96.78 ($\pm$ 0.57)} & \underline{60.15 ($\pm$ 1.82)} & 91.38 ($\pm$ 0.42) & \textbf{97.57 ($\pm$ 0.37)} &  9.77 ($\pm$ 0.79) & 88.92 ($\pm$ 0.72) \\
     &Reverse Distillation \cite{deng2022anomaly} &  87.19 ($\pm$ 0.99) & 57.22 ($\pm$ 5.77) & 84.04 ($\pm$ 1.64) & 86.09 ($\pm$ 1.34) & 10.64 ($\pm$ 1.55) & 78.83 ($\pm$ 2.66) \\
   \rowcolor{tabBlue} & \ourmethod{} & 94.78 ($\pm$ 1.09) & \textbf{85.81 ($\pm$ 1.23)} & \textbf{92.53 ($\pm$ 0.81)} & \underline{96.92 ($\pm$ 0.57)} & \textbf{28.73 ($\pm$ 1.70)} & \textbf{92.97 ($\pm$ 0.36)} \\
    \bottomrule
   \end{tabular}
   }
\end{table*}

\subsection{\sukanya{Image Decoder}}
\label{sec:dec}

To predict the \(i\)-th image, we use \(c_{i}\), the past context encoding, \(\psi_{i}\), the embedding of the next inter-arrival time \(\tau_{i}\), as well as \(\Psi_{i}\), the embedding of the time duration since the start of the operation \(\delta_{i}\). Specifically, we compute \sukanya{\(\hat{x}_{i} = \phi_d([c_{i} \oplus (\psi_{i} + \Psi_{i})]; W_d)\)} where \(\phi_d(\cdot; W_d)\) is the decoder network. Note that the decoder network is pre-trained along with the image encoder using the image reconstruction task on the images from the training set \(\mathcal{D}_N\). \sukanya{The prediction error is computed as the Frobenius norm difference between the original and the forecasted image. The total training loss is obtained by averaging the prediction errors over all the training examples:}
\begin{equation}
    \label{eq:loss}
    \sukanya{\mathcal{L}_{\text{train}} = \frac{1}{|\mathcal{D}_N|} \sum_{i=1}^{|\mathcal{D}_N|} \|x_{i} - \hat{x}_{i} \|_{F}^2} 
\end{equation}
Finally, \sukanya{the anomaly score of a new point $(x, y, t)$ is defined as the associated prediction error, i.e. $s(x, t) = \|x - \hat{x} \|_{F}^2$}. Algorithm~\ref{algo:forecastAD} summarizes the different steps of our \ourmethod{} method.
\IncMargin{1.5em}
\begin{algorithm}[ht]
\SetAlgoLined
\DontPrintSemicolon
\SetNoFillComment
\Indm
\KwInput{\sukanya{$\textup{Training dataset}~\mathcal{D}_N,\, \textup{Sinusoidal encoder}~f_{\mathrm{sin}}$ \\ $\textup{Image encoder}~\phi_e,\, \textup{Image decoder}~\phi_d,\, \textup{Number of epochs}~e$ \\ $\textup{Learning rate}~\eta,\, \textup{Context length}~K$}}
\Indp
        \For{\sukanya{$(\textup{epoch} = 1, 2, \cdots,e)~\textup{and}~((x_i, t_i, y_i) \in \mathcal{D}_N)$}}{

            {Initialize context embedding:}\;
            \quad $c_{i} \leftarrow \texttt{random()}$\\
            \For{$k = K, K-1, \cdots, 1$}{
                {Calculate the time embeddings:}\;
                \quad \sukanya{$\psi_{i-k} \leftarrow f_{\mathrm{sin}}(\tau_{i-k})$}\; 
                \quad \sukanya{$\Psi_{i-k} \leftarrow f_{\mathrm{sin}}(\delta_{i-k})$}\\
                {Create joint embedding:}\;
                \quad \sukanya{$\hat{z}_{i-k} \leftarrow \left[\phi_e(x_{i-k}; W_e)\oplus\; (\psi_{i-k} + \Psi_{i-k})\right]$}\\
                {Update context embedding}\;
                \quad $c_{i} \leftarrow \varphi(c_{i}, \hat{z}_{i-k}; W_c)$\\
            }
            {Encode target time embeddings:}\;
            \quad \sukanya{$\psi_{i} \leftarrow f_{\mathrm{sin}}(\tau_{i}),\, \Psi_{i} \leftarrow f_{\mathrm{sin}}(\delta_{i})$}\\
            {Predict the next data point:}\;
            \quad \sukanya{$\hat{x}_{i} \leftarrow \phi_d([c_i \oplus (\psi_{i} + \Psi_{i})]; W_d)$}\\
            {\sukanya{Update the model parameters $W_e$, $W_d$ and $W_c$ by minimising the loss $\mathcal{L}_{\text{train}}$ (Eq.~\ref{eq:loss})}}
        }
\Indm
\KwOutput{$\phi_e(\cdot;W_e),\, \phi_d(\cdot;W_d),\, \varphi(\cdot;W_c)$}
\Indp
\caption{\sukanya{Training process of \ourmethod{}}}
\label{algo:forecastAD}
\end{algorithm}
\DecMargin{1.5em}

\section{Experiments}
\label{sec:eval}

\subsection{Baselines}

We \sukanya{first} compare \ourmethod{} against \sukanya{simple methods, which detect anomalies based on statistical features extracted from the images}. These features include the corresponding time of day, as well as the mean, maximum, and standard deviation of the temperature, to distinguish between normal and abnormal samples. We also evaluate against deep image-based AD methods, namely, autoencoder, FastFlow \cite{Yu2021FastFlow:Flows}, PatchCore \cite{Roth2022TowardsDetection}, PaDiM \cite{Defard2021PaDiM:Localization}, CFlow \cite{Gudovskiy2021CFLOW-AD:Flows}, DR{\AE}M \cite{Zavrtanik2021DRMDetection}, and Reverse Distillation \cite{deng2022anomaly}. Deep methods have been shown to be more effective than shallow ones for image AD \cite{Ruff2021ADetection}, leveraging the deep neural networks' capability to extract representative features through multiple layers of abstraction.

\subsection{Experimental Setup}
\textbf{Network architectures and hyperparameters}. Except for the autoencoder, the baselines follow the implementation from Anomalib \cite{Akcay2022Anomalib:Detection}, which is a widely used library for benchmarking AD methods. Based on our experiments, we opted for a Deep Convolutional Autoencoder (DCAE) with a latent dimension of \sukanya{$d' = 128$}. Detailed architectural specifications are provided in Appendix~\ref{sec:arch}. The image encoder employed in \ourmethod{} mirrors the structure of the downsampling branch in DCAE. In \ourmethod{}, we adopt a \sukanya{$4$-layer} LSTM network with a hidden dimension of \sukanya{$128$} to serve as the context encoder $\varphi$. For time encoding, the sinusoidal embedding has a dimension of $16$. We adhere to the hyperparameters mentioned by the authors for the baseline methods. For \ourmethod{}, we use MSE and train using an Adam optimizer with a learning rate of $0.001$ and weight decay of $0.00001$. We use a pre-processing step for all the experiments where the images in the dataset are resized to $256 \times 256$ to be compatible with the baselines. Unless otherwise specified, we use a sequence length of $K=30$.

\noindent \textbf{Dataset}. Our labelled dataset comprises days, which are segmented into training, validation, and test sets. Days featuring exclusively normal samples are allocated across these three sets, while those with anomalous samples are included in both the validation and test sets. To underscore the challenges presented by low-temperature samples, we adopt two training setups: (i) \textbf{[Tr\#1]}, incorporating training and validation samples solely from the M phase, and (ii) \textbf{[Tr\#2]}, comprising training and validation samples from the S, M, and E phases. Importantly, the test set in both scenarios consists of samples spanning the S, M, and E phases. The distribution of normal and anomalous samples across S, M, and E phases for these setups is depicted in Figure~\ref{fig:training_setup}. For \ourmethod{}, we generate a sequence for each \sukanya{data point} by selecting \(K\) preceding samples. \sukanya{If there are less than \(K\) prior samples, we duplicate the corresponding day's first data point to form a \(K\)-length sequence. Lastly, for the first data point captured each day, we set the \(\tau\) and \(\delta\) to a small positive value \(\epsilon = 1e-5\).}

\noindent \textbf{Model evaluation}. We evaluate the models based on the Area under the Receiver Operating Characteristics curve (AUROC) and the Area under the Precision-Recall curve (AUPR). To highlight the effectiveness of each model in distinguishing between low-temperature normal and anomalous behaviours, we utilize three test setups containing: (i) test samples in M \textbf{[Ts\#1]}, (ii) test samples in S-E \textbf{[Ts\#2]}, and (iii) test samples in S-M-E \textbf{[Ts\#3]}. For the experiments below, we report mean over 5 runs along with one standard error.

\subsection{Results and Discussion}

We summarize the results over five runs for different training setups in Table \ref{tab:res}. For the training setup \textbf{[Tr\#1]}, \ourmethod{} provides competitive results when compared to image-based SOTA models as measured by both AUROC and AUPR metrics over the test samples in \textbf{[Ts\#3]}. Additionally, we observe good performance for image-based SOTA approaches in \textbf{[Ts\#1]}. This performance can be attributed to the training exclusively on samples from M, which predominantly fall within the high-temperature region where temporal context is less critical. \sukanya{However, since the models are not trained on low-temperature normal samples from the start and end of the operational cycle, their performance in \textbf{[Ts\#2]} naturally declines. Specifically, the AUPR score is significantly low in \textbf{[Ts\#2]}, as the models tend to assign very high anomaly scores to most low-temperature samples found in S-E.}

Considering the setup \textbf{[Tr\#2]}, we observe a drop in performance for the SOTA methods compared to \textbf{[Tr\#1]}. \sukanya{This observation can be attributed to the fact that when the model is exposed to a limited number of low-temperature normal samples, it struggles to learn from them.} Instead, these samples act as contamination, diminishing performance over the high-temperature samples in \textbf{[Ts\#1]}. Additionally, baselines fail to distinguish between low-temperature normal and anomalous samples in \textbf{[Ts\#2]}, as they do not incorporate temporal features. \ourmethod{} significantly outperforms all baselines by approx. 25\% in \textbf{[Ts\#2]}, while maintaining competitive performance across all test samples in \textbf{[Ts\#3]}. \sukanya{In Appendix \ref{sec:extended_res}, we provide an extended version of Table \ref{tab:res}.}

\subsection{Ablation Study}

\noindent \sukanya{\textbf{Importance of time-embedding and pre-training}. Table~\ref{tab:ab1} shows the results of an ablation study to understand the importance of $\tau$ and $\delta$ in \ourmethod{}. In all configurations, we always keep the image encoding as part of the input. Firstly, we observe the lowest AUROC and AUPR scores in \textbf{[Ts\#2]} when the context has only the encodings of $K$-prior images. It emphasizes the need to address the challenge posed by irregular sequences and co-occurrence of low-temperature normal and anomalous samples. Then, on considering either $\tau$ or $\delta$, we observe a significant improvement in \textbf{[Ts\#2]}. Furthermore, incorporating both $\tau$ and $\delta$ yields the best performance, highlighting that both $\tau$ and $\delta$ are necessary for reliable detection of anomalies. Lastly, we also empirically validate the impact of pre-training the image encoder and decoder using the image reconstruction task. Using the pre-trained models offers substantial enhancements in performance when compared to a randomly-initialized backbone.}

\begin{table}[h]
  \caption{\sukanya{Ablation of time-embedding and pre-training.}}\vspace{-1em}
  \label{tab:ab1}
  \centering
  \aboverulesep = 0pt 
  \belowrulesep = 0pt
  \renewcommand{\arraystretch}{1.2}
  \resizebox{\columnwidth}{!}{
  \scriptsize
  \begin{tabular}{@{}c:cc|ccc|ccc@{}}
    \toprule
    \multirow{2}{*}{\textbf{Pre-train}} & \multirow{2}{*}{$\tau$} & \multirow{2}{*}{$\delta$} & \multicolumn{3}{c|}{\textbf{AUROC (\%)}} & \multicolumn{3}{c}{\textbf{AUPR (\%)}}\\
    &  &  & \textbf{[Ts\#1]} & \textbf{[Ts\#2]} & \textbf{[Ts\#3]} & \textbf{[Ts\#1]} & \textbf{[Ts\#2]} & \textbf{[Ts\#3]}\\
    \midrule
    - & \checkmark & \checkmark & 94.60 ($\pm$ 1.60) & 75.30 ($\pm$ 5.89) & 90.58 ($\pm$ 1.00) & \underline{96.78 ($\pm$ 0.81)} & 23.54 ($\pm$ 4.17) & 89.93 ($\pm$ 1.29) \\
    \cdashline{1-9}
    \checkmark & - & - & \textbf{97.12 ($\pm$ 0.44)} & 72.94 ($\pm$ 7.15) & \underline{92.74 ($\pm$ 1.26)} & 98.06 ($\pm$ 0.29) & 21.32 ($\pm$ 4.29) & 91.41 ($\pm$ 1.88) \\
    \checkmark & \checkmark & - & 94.59 ($\pm$ 0.93) & \underline{84.08 ($\pm$ 3.83)} & 92.49 ($\pm$ 0.79) & 96.56 ($\pm$ 0.49) & \underline{28.83 ($\pm$ 4.17)} & \underline{92.67 ($\pm$ 0.57)} \\
    \checkmark & - & \checkmark & 92.71 ($\pm$ 1.32) & 82.71 ($\pm$ 3.09) & 91.12 ($\pm$ 1.09) & 95.58 ($\pm$ 0.95) & 26.31 ($\pm$ 3.60) & 92.15 ($\pm$ 0.96) \\
    \rowcolor{tabBlue}\checkmark & \checkmark & \checkmark & \underline{94.78 ($\pm$ 1.09)} & \textbf{85.81 ($\pm$ 1.23)} & \textbf{92.53 ($\pm$ 0.81)} & \textbf{96.92 ($\pm$ 0.57)} & \textbf{28.73 ($\pm$ 1.70)} & \textbf{92.97 ($\pm$ 0.36)} \\
    \bottomrule
  \end{tabular}
  }
\end{table}

\noindent \textbf{Effect of context length ($K$)}. We report the AD performance of \ourmethod{} with varying context lengths $K$ in Table \ref{tab:ab2}.  For context lengths $K \leq 20$, we do not observe any correlation between performance and context length. However, larger sequence lengths of 30 or 40 yield better performance. To limit computational demands, we did not consider larger sequence lengths and chose a sequence length of $30$ for all our experiments. \\

\begin{table}[h]
  \caption{\sukanya{Ablation of K}}\vspace{-1em}
  \label{tab:ab2}
  \centering
  \aboverulesep = 0pt 
  \belowrulesep = 0pt
  \renewcommand{\arraystretch}{1.2}
  \resizebox{\columnwidth}{!}{
  \scriptsize
  \begin{tabular}{@{}c|ccc|ccc@{}}
    \toprule
    \multirow{2}{*}{\textbf{K}} & \multicolumn{3}{c|}{\textbf{AUROC (\%)}} & \multicolumn{3}{c}{\textbf{AUPR (\%)}}\\
     & \textbf{[Ts\#1]} & \textbf{[Ts\#2]} & \textbf{[Ts\#3]} & \textbf{[Ts\#1]} & \textbf{[Ts\#2]} & \textbf{[Ts\#3]}\\
    \midrule
    1 & 88.85 ($\pm$ 2.55) & 78.26 ($\pm$ 1.86) & 83.64 ($\pm$ 1.72) & 92.68 ($\pm$ 1.49) & 23.14 ($\pm$ 2.63) & 83.22 ($\pm$ 1.01) \\
    5 & 91.21 ($\pm$ 0.94) & \textbf{87.83 ($\pm$ 1.45)} & 89.25 ($\pm$ 0.81) & 94.44 ($\pm$ 0.51) & \textbf{32.24 ($\pm$ 2.15)} & 89.82 ($\pm$ 0.49) \\
    10 & \underline{94.02 ($\pm$ 1.81)} & 78.13 ($\pm$ 3.91) & 89.82 ($\pm$ 0.77) & \underline{96.40 ($\pm$ 0.93)} & 23.05 ($\pm$ 2.07) & 89.29 ($\pm$ 0.86) \\
    20 & 92.64 ($\pm$ 1.21) & 83.22 ($\pm$ 2.90) & 90.46 ($\pm$ 1.31) & 95.65 ($\pm$ 0.60) & 27.20 ($\pm$ 3.45) & 91.39 ($\pm$ 0.93) \\
    \rowcolor{tabBlue}30 & \textbf{94.78 ($\pm$ 1.09)} & 85.81 ($\pm$ 1.23) & \textbf{92.53 ($\pm$ 0.81)} & \textbf{96.92 ($\pm$ 0.57)} & 28.73 ($\pm$ 1.70) & \textbf{92.97 ($\pm$ 0.36)} \\
    40 & 92.66 ($\pm$ 1.46) & 85.87 ($\pm$ 0.92) & 91.13 ($\pm$ 1.26) & 95.59 ($\pm$ 0.73) & 29.24 ($\pm$ 1.49) & 91.88 ($\pm$ 0.83) \\
    50 &  93.44 ($\pm$ 0.72) & \underline{87.29 ($\pm$ 1.77)} & \underline{92.09 ($\pm$ 0.45)} & 96.06 ($\pm$ 0.34) & \underline{31.51 ($\pm$ 1.52)} & \underline{92.62 ($\pm$ 0.31)} \\
    60 & 93.36 ($\pm$ 0.75) & 81.03 ($\pm$ 4.25) & 91.03 ($\pm$ 1.51) & 95.95 ($\pm$ 0.44) & 28.46 ($\pm$ 3.83) & 91.45 ($\pm$ 1.58) \\
    \bottomrule
  \end{tabular}
  }
\end{table}

\begin{figure}[h]
    \centering
    \includegraphics[width=\columnwidth]{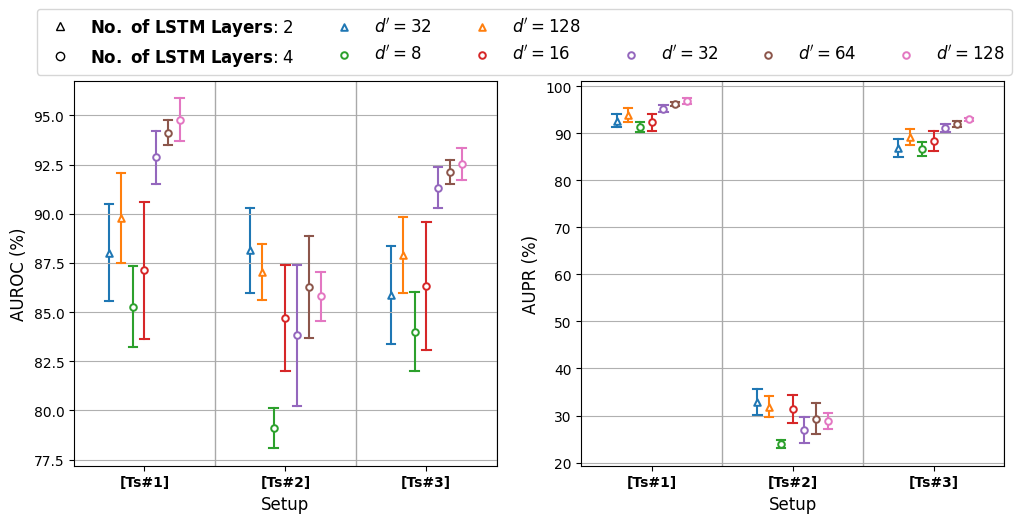}
    \vspace{-2em}
    \caption{Ablation of different architectures}
    \label{fig:abla_arch}
\end{figure}

\noindent \sukanya{\textbf{Effect of different architecture}. In Figure~\ref{fig:abla_arch}, we analyze the effect of the number of layers in LSTM and the latent dimension \(d'\) on the AUROC and AUPR scores. Firstly, we observe that larger latent dimensions lead to higher scores in most cases, regardless of the number of layers in LSTM. Secondly, \ourmethod{} performs better on \textbf{[Ts\#1]} and \textbf{[Ts\#3]} with a $4$-layer LSTM, while a $2$-layer LSTM yields better results on \textbf{[Ts\#2]}. Based on this empirical observation, we chose a $4$-layer LSTM with latent dimension \(d'=128\), which has the highest scores in \textbf{[Ts\#1]} and \textbf{[Ts\#3]} while having a comparable performance with the best configuration on \textbf{[Ts\#2]}.}

\subsection{\sukanya{Interpretability of \ourmethod{}}}
    \begin{figure*}[t]
        \begin{subfigure}{0.24\textwidth}
            \centering
            \includegraphics[width=\textwidth]{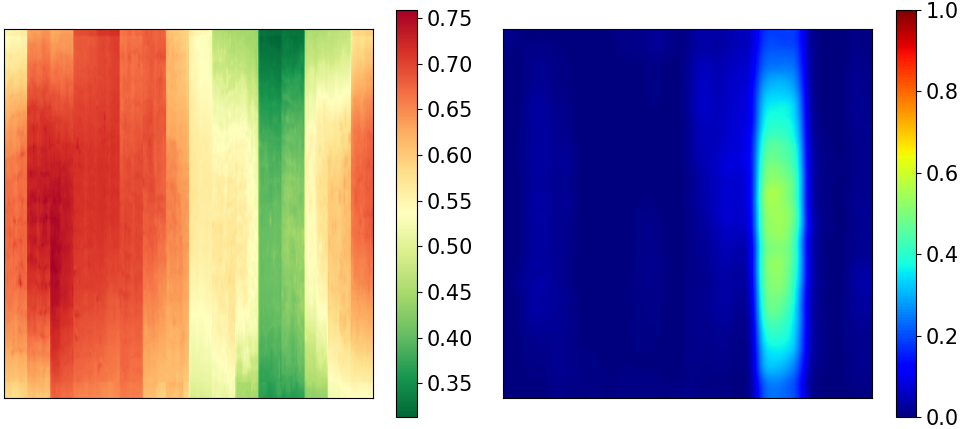}
        \end{subfigure}
        \begin{subfigure}{0.24\textwidth}
            \centering
            \includegraphics[width=\textwidth]{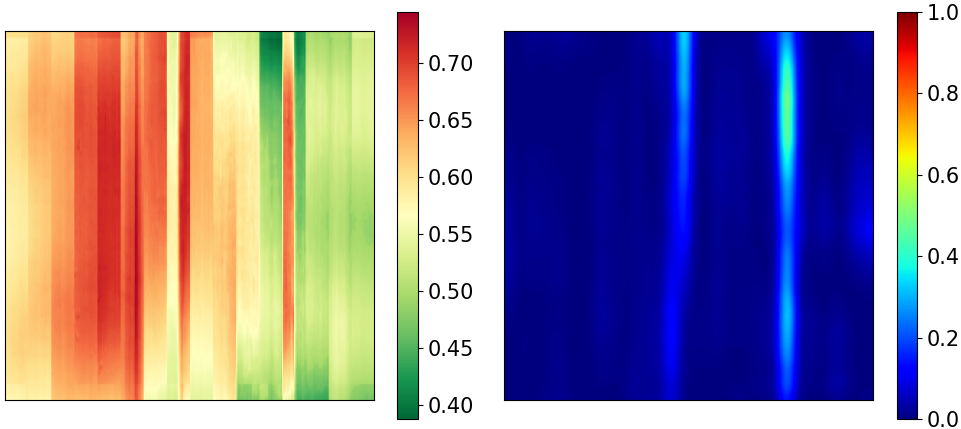}
        \end{subfigure}
        \begin{subfigure}{0.24\textwidth}
            \centering
            \includegraphics[width=\textwidth]{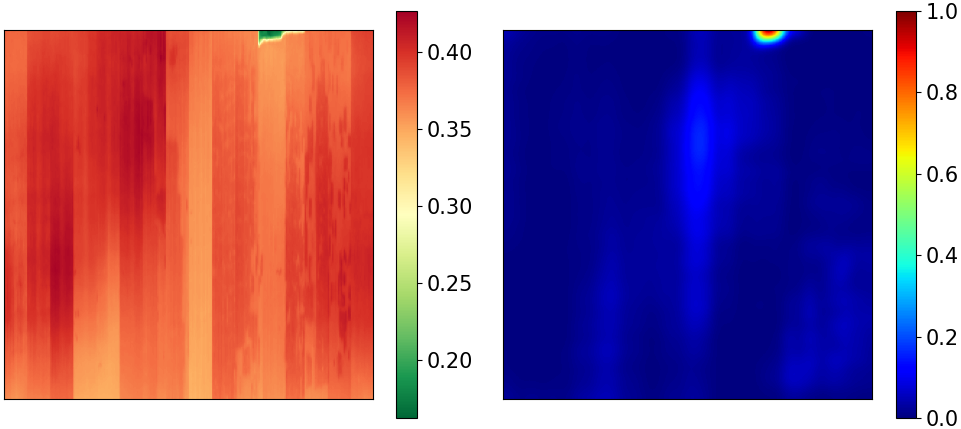}
        \end{subfigure}
        \begin{subfigure}{0.24\textwidth}
            \centering
            \includegraphics[width=\textwidth]{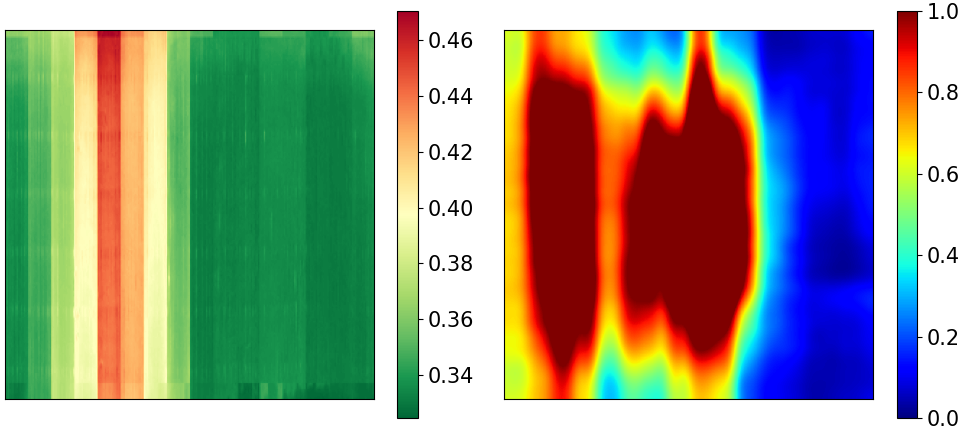}
        \end{subfigure}

        \begin{subfigure}{0.24\textwidth}
            \centering
            \includegraphics[width=\textwidth]{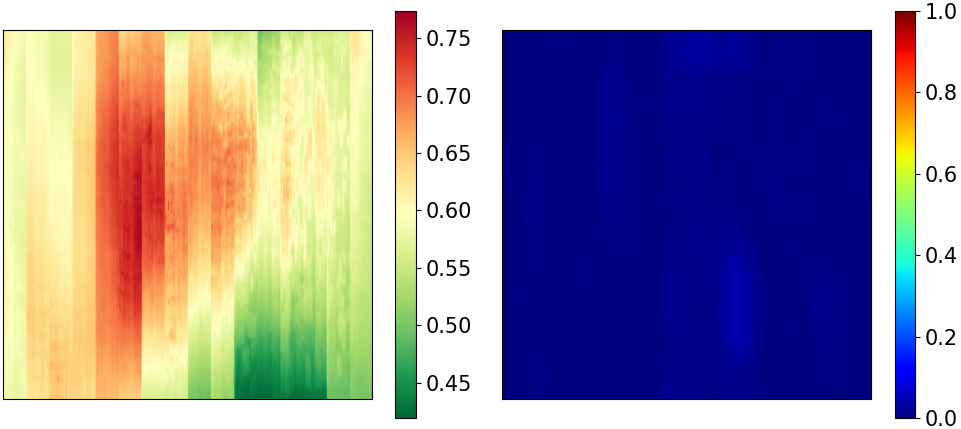}
        \end{subfigure}
        \begin{subfigure}{0.24\textwidth}
            \centering
            \includegraphics[width=\textwidth]{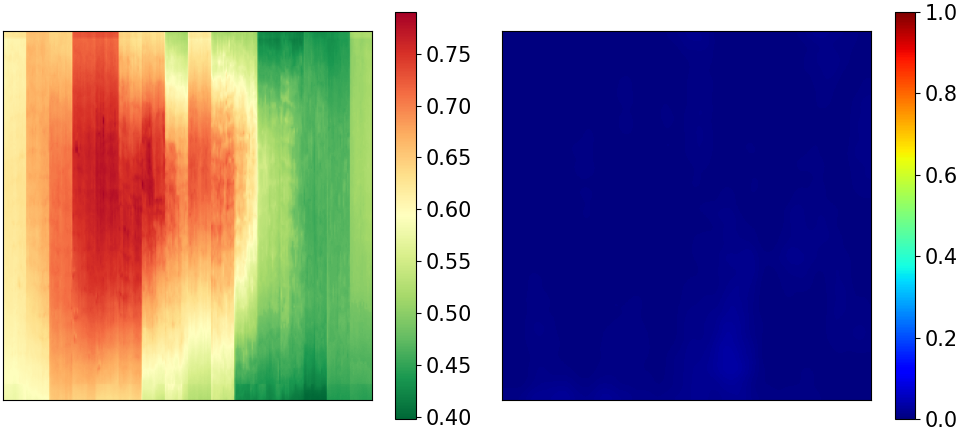}
        \end{subfigure}
        \begin{subfigure}{0.24\textwidth}
            \centering
            \includegraphics[width=\textwidth]{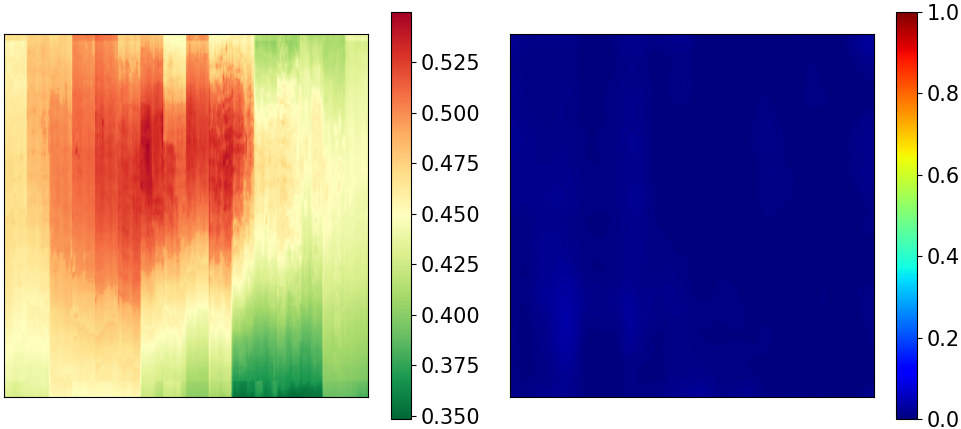}
        \end{subfigure}
        \begin{subfigure}{0.24\textwidth}
            \centering
            \includegraphics[width=\textwidth]{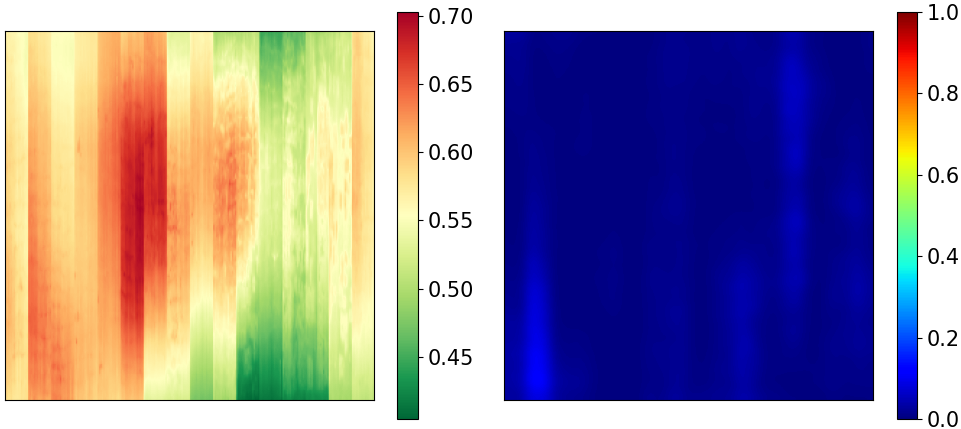}
        \end{subfigure}
        \vspace{-1em}
        \caption{\sukanya{Examples of anomaly maps for anomalous (top) and normal (bottom) images.}}
        \label{fig:maps}
    \end{figure*}

\sukanya{Interpretability of deep learning models is critical for high-risk applications to enhance transparency and trustworthiness. Therefore, we extract anomaly maps from \ourmethod{} corresponding to each image during inference. Recall that \ourmethod{} is trained with pixel-wise regression loss, and thus, the anomaly map can be computed as the difference between the original and forecasted images. Based on recent works on IAD \citep{Roth2022TowardsDetection, Defard2021PaDiM:Localization}, we smoothed the anomaly maps using a Gaussian filter and normalized it using the minimum and maximum anomaly scores for the normal samples in the validation set. In Figure~\ref{fig:maps}, we show the anomaly maps of 4 normal and 4 anomalous test samples, along with the image for reference. It can be seen that for specific types of anomalies, such as freezing, where we observe high-temperature streaks, \ourmethod{} assign high anomaly scores to those regions. Therefore, it aids the interpretability of the results from \ourmethod{}. To further enhance the understanding, the anomaly maps can be complemented by plots of mean temperature to show the sudden drops or rises in temperature resulting in the samples being anomalous.}

\subsection{\sukanya{Simulated Dataset}}
\begin{figure}[h]
    \centering
    \includegraphics[width=\columnwidth]{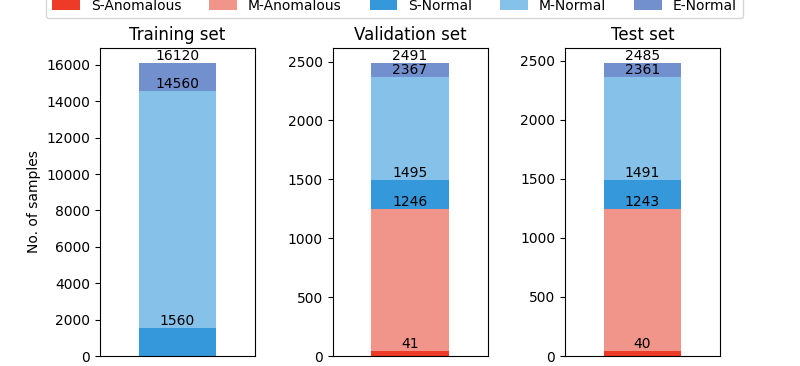}
    \vspace{-1em}
    \caption{Data split for simulated dataset}
    \label{fig:sim_split}
\end{figure}

\sukanya{We have prepared a simulated dataset to ensure reproducibility and validation of the results. We use a variational autoencoder to generate the data. Additional details about the data generation are deferred to Appendix~\ref{sec:datagen}. The distribution of normal and anomalous samples across S, M, and E phases for these setups is depicted in Figure~\ref{fig:sim_split}. We have also compared our method to the baselines on the simulated dataset. The results are reported in Table~\ref{tab:sim_res} for training setup \textbf{[Tr\#2]} and test setup \textbf{[Ts\#3]}, which are the main focus of our work. The results highlight the effectiveness of \ourmethod{}, similar to our results on the original dataset as reported in Table~\ref{tab:res} of the paper. Furthermore, in Appendix~\ref{sec:datagen}, we provide qualitative evidence to support the validity of the simulated dataset by visualizing generated and original images for a random set of timestamps.}

\begin{table}[h]
    \centering
    \caption{\sukanya{Anomaly detection performance on simulated data.}}\vspace{-1em}
    \label{tab:sim_res}
    \aboverulesep = 0pt
    \belowrulesep = 0pt
    \renewcommand{\arraystretch}{1.2}
    \resizebox{\linewidth}{!}{
    \begin{tabular}{@{}l|ccc|ccc@{}}
    \toprule
    \multirow{2}{*}{\textbf{Model}} &  \multicolumn{3}{c|}{\textbf{AUROC} (\%)} &  \multicolumn{3}{c}{\textbf{AUPR} (\%)}\\
    &  \textbf{[Ts\#1]} & \textbf{[Ts\#2]} & \textbf{[Ts\#3]} & \textbf{[Ts\#1]} & \textbf{[Ts\#2]} & \textbf{[Ts\#3]}\\
    \midrule 
    Autoencoder & 87.97 ($\pm$ 4.08) & 66.34 ($\pm$ 2.49) & 82.00 ($\pm$ 1.58) & 94.04 ($\pm$ 1.99) & 24.72 ($\pm$ 6.51) & 83.46 ($\pm$ 1.61) \\

     CFlow \cite{Gudovskiy2021CFLOW-AD:Flows} & 83.42 ($\pm$ 2.97) & 51.32 ($\pm$ 4.16) & 70.30 ($\pm$ 2.67) & 90.67 ($\pm$ 1.97) & 10.46 ($\pm$ 0.80) & 69.42 ($\pm$ 2.14) \\

     DR{\AE}M \cite{Zavrtanik2021DRMDetection}& 98.11 ($\pm$ 0.81) & 61.89 ($\pm$ 5.32) & 89.02 ($\pm$ 0.81) & \underline{99.02 ($\pm$ 0.40)} & \underline{25.90 ($\pm$ 4.32)} & 88.52 ($\pm$ 0.75) \\

     FastFlow \cite{Yu2021FastFlow:Flows} & 97.24 ($\pm$ 0.54) & 52.23 ($\pm$ 3.63) & 87.98 ($\pm$ 0.67) & 98.43 ($\pm$ 0.26) &  9.49 ($\pm$ 0.63) & 87.76 ($\pm$ 0.93) \\

     PaDiM  \cite{Defard2021PaDiM:Localization}& 97.93 ($\pm$ 0.56) & 56.04 ($\pm$ 0.42) & 88.97 ($\pm$ 0.44) & 98.76 ($\pm$ 0.31) &  9.89 ($\pm$ 0.07) & 88.25 ($\pm$ 0.25) \\

     PatchCore \cite{Roth2022TowardsDetection}& \underline{98.28 ($\pm$ 0.29)} & \underline{66.42 ($\pm$ 1.96)} & \underline{92.31 ($\pm$ 0.31)} & 98.81 ($\pm$ 0.20) & 21.57 ($\pm$ 2.79) & \underline{92.28 ($\pm$ 0.26)} \\

     Reverse Distillation \cite{deng2022anomaly}& 75.80 ($\pm$ 5.53) & 57.59 ($\pm$ 4.25) & 65.60 ($\pm$ 4.60) & 86.23 ($\pm$ 3.10) & 12.14 ($\pm$ 1.95) & 63.36 ($\pm$ 3.35) \\
   \rowcolor{tabBlue}\ourmethod{} & \textbf{98.73 ($\pm$ 0.64)} & \textbf{98.61 ($\pm$ 0.43)} & \textbf{97.84 ($\pm$ 0.65)} & \textbf{99.30 ($\pm$ 0.33)} & \textbf{88.03 ($\pm$ 3.18)} & \textbf{97.96 ($\pm$ 0.54)} \\
    \bottomrule
   \end{tabular}
   }
\end{table}

\subsection{Deployment}

\sukanya{We have tested \ourmethod{} over five months of data from an operational CSP plant. A freshly labelled dataset was curated by initially applying a predefined set of labelling rules, followed by a meticulous review and cleanup of the dataset with guidance from domain experts. The performance metrics of \ourmethod{} on this labelled set containing 8373 normal and 1,321 abnormal samples are detailed in Table~\ref{tab:deployment_monthly}. Furthermore, the deployment results are broken down per month over different operating stages. Please note that for some months, we could not compute the performance metrics as there are no anomalous samples present in the dataset. Such cases are marked as ``$-$'' in the table. It is important to note that there is variability in this data, such as different stages of operations (starting, ending, and middle) and varying external weather conditions. We can observe that \ourmethod{} is fairly robust in the detection of anomalies over this period. The actionable insights derived from \ourmethod{} contribute to the strategic maintenance planning of the CSP plant, thereby enhancing the durability of its equipment.}

\begin{table}[h]
  \caption{\sukanya{Deployment performance}}\vspace{-1em}
  \label{tab:deployment_monthly}
  \centering
  \aboverulesep = 0pt 
  \belowrulesep = 0pt
  \renewcommand{\arraystretch}{1.2}
  \resizebox{0.8\columnwidth}{!}{
  \footnotesize
  \begin{tabular}{@{}c|ccc|ccc@{}}
    \toprule
    \multirow{2}{*}{\textbf{Month}} & \multicolumn{3}{c|}{\textbf{AUROC (\%)}} & \multicolumn{3}{c}{\textbf{AUPR (\%)}}\\
     & \textbf{[Ts\#1]} & \textbf{[Ts\#2]} & \textbf{[Ts\#3]} & \textbf{[Ts\#1]} & \textbf{[Ts\#2]} & \textbf{[Ts\#3]}\\
    \midrule
    1 &       0.89 &           0.71 &         0.89 &      0.66 &           0.17 &        0.62 \\
    2 &       0.86 &           0.94 &         0.85 &      0.82 &           0.70 &        0.79 \\
    3 &       0.96 &           0.93 &         0.95 &      0.91 &           0.19 &        0.87 \\
    4 &       0.91 & - &         0.91 &      0.75 & - &        0.63 \\
    5 &       0.85 & - &         0.85 &      0.60 & - &        0.57 \\
    \midrule
    Overall & 0.88 &        0.81 &         0.88 &      0.72 &       0.25 &        0.69 \\
    \bottomrule
    \end{tabular}
  }
\end{table}

\section{Conclusion}
\label{sec:conclusion}
We address the problem of anomaly detection in irregular sequences of thermal images collected from IR cameras in an operational CSP plant. Extensive analysis of our dataset reveals distinctive temporal characteristics, setting it apart from established AD industrial image benchmark datasets like MVTec \cite{Bergmann2019MVTECDetection}. We empirically demonstrate that image-based SOTA AD methods underperform, especially when context is critical for anomaly detection. We also introduce a forecasting-based AD method, \ourmethod{}, that predicts future thermal images from past sequences and timestamps using a deep sequence model. This method effectively captures specific temporal data features and distinguishes between difficult-to-detect temperature-based anomalies. Experimental results demonstrate the effectiveness of \ourmethod{}, outperforming existing SOTA methods as measured by AUROC and AUPR. Notably, \ourmethod{} exhibits significant enhancements in detecting anomalous behaviours, particularly among low-temperature samples. Furthermore, \ourmethod{} has been successfully deployed, providing critical insights for the maintenance of the CSP plant to our industry partner. For future work, we aim to further study the role of context and sequence lengths in anomaly detection performance. We also aim to extend our model to be more robust to distribution shifts inherent in industrial processes, notably by considering probabilistic forecasting models.   

\begin{acks}
This work is supported by the research project ``Federated Learning and Augmented Reality for Advanced Control Centers''. We thank \sukanya{Thibault GEORGES and Adrien FARINELLE} from John Cockerill for helping us understand the dataset along with the associated abnormal behaviours.
\end{acks}

\clearpage

\bibliographystyle{ACM-Reference-Format}
\bibliography{references, biblio}

\clearpage
\appendix

\section{Network Architecture}
\label{sec:arch}
\ourmethod{} is implemented using the PyTorch framework \cite{Paszke2019PyTorch:Library}. The architecture of the encoder $\phi_e$ and $\phi_d$ is presented in Tables~\ref{tab:enc} and \ref{tab:dec}, respectively. The input dimension of the encoder network is (3 $\times$ 256 $\times$ 256). The period of the sinusoidal encoding is 1000.
\begin{table}[ht]
    \centering
    \caption{Encoder architecture}\vspace{-1em}
    \label{tab:enc}
    \aboverulesep = 0pt
    \belowrulesep = 0pt
    \renewcommand{\arraystretch}{1.2}
    \resizebox{0.7\linewidth}{!}{
    \begin{tabular}{@{}l@{}}
    \toprule
    \textbf{Layer}\\
    \midrule 
    Conv2d-1(3, 32, 5, padding=2, stride=2) \\
    BatchNorm2d-1(32, eps=1e-04)\\
    MaxPool2d-1(2,2)\\
    Conv2d-2(32, 64, 5, padding=2, stride=2) \\
    BatchNorm2d-2(64, eps=1e-04)\\
    MaxPool2d-2(2,2)\\
    Conv2d-3(64, 128, 5, padding=2, stride=2) \\
    BatchNorm2d-3(128, eps=1e-04)\\
    MaxPool2d-3(2,2)\\
    Conv2d-4(128, 128, 5, padding=2, stride=2) \\
    BatchNorm2d-4(128, eps=1e-04)\\
    MaxPool2d-4(2,2)\\
    Linear-1(128, 128, padding=2, stride=2) \\
    BatchNorm2d-1(128, eps=1e-04)\\
    \bottomrule
   \end{tabular}
   }
\end{table}

\begin{table}[ht]
    \centering
    \caption{Decoder architecture}\vspace{-1em}
    \label{tab:dec}
    \aboverulesep = 0pt
    \belowrulesep = 0pt
    \renewcommand{\arraystretch}{1.2}
    \resizebox{0.8\linewidth}{!}{
    \begin{tabular}{@{}l@{}}
    \toprule
    \textbf{Layer}\\
    \midrule 
    ConvTranspose2d-1(128, 64, 5, padding=2) \\
    BatchNorm2d-1(64, eps=1e-04)\\
    ConvTranspose2d-2(64, 64, 5, padding=2) \\
    BatchNorm2d-2(64, eps=1e-04)\\
    ConvTranspose2d-3(64, 32, 5, padding=2) \\
    BatchNorm2d-3(32, eps=1e-04)\\
    ConvTranspose2d-4(32, 3, 5, padding=2) \\
    \bottomrule
   \end{tabular}
   }
\end{table}

\section{Data generation }
\label{sec:datagen}

\sukanya{We generate a public dataset mirroring the properties of our private dataset using a variational autoencoder (VAE). For daily sequence generation, we condition the VAE on the time of day, class (Positive, Negative, Unlabelled), and phase (Start, Middle, End). A standard convolutional neural network (CNN) with three convolution blocks with an increasing number of feature maps is used to encode images, and a multilayer perceptron (MLP) is used to encode conditioning variables. Then, a two-layer LSTM network produces the context embedding. Finally, a deconvolutional CNN, mirroring the encoder's architecture, reconstructs the original image from a latent vector sampled from the latent distribution. We use a multivariate Gaussian distribution with a diagonal covariance matrix as our latent distribution. We train the VAE over 20 epochs using Adam optimizer, a learning rate of 1e-4, and a batch size of 16. Figure~\ref{fig:samples_comparison} allows us to compare the generated images with the original images from the private dataset. From the figure, we can observe that the VAE can generate normal images and a diverse set of anomalies which are visually similar to images in the original dataset.}

\section{Sensitivity to Data Labelling}

\begin{figure*}[h]
   \includegraphics[width=\linewidth]{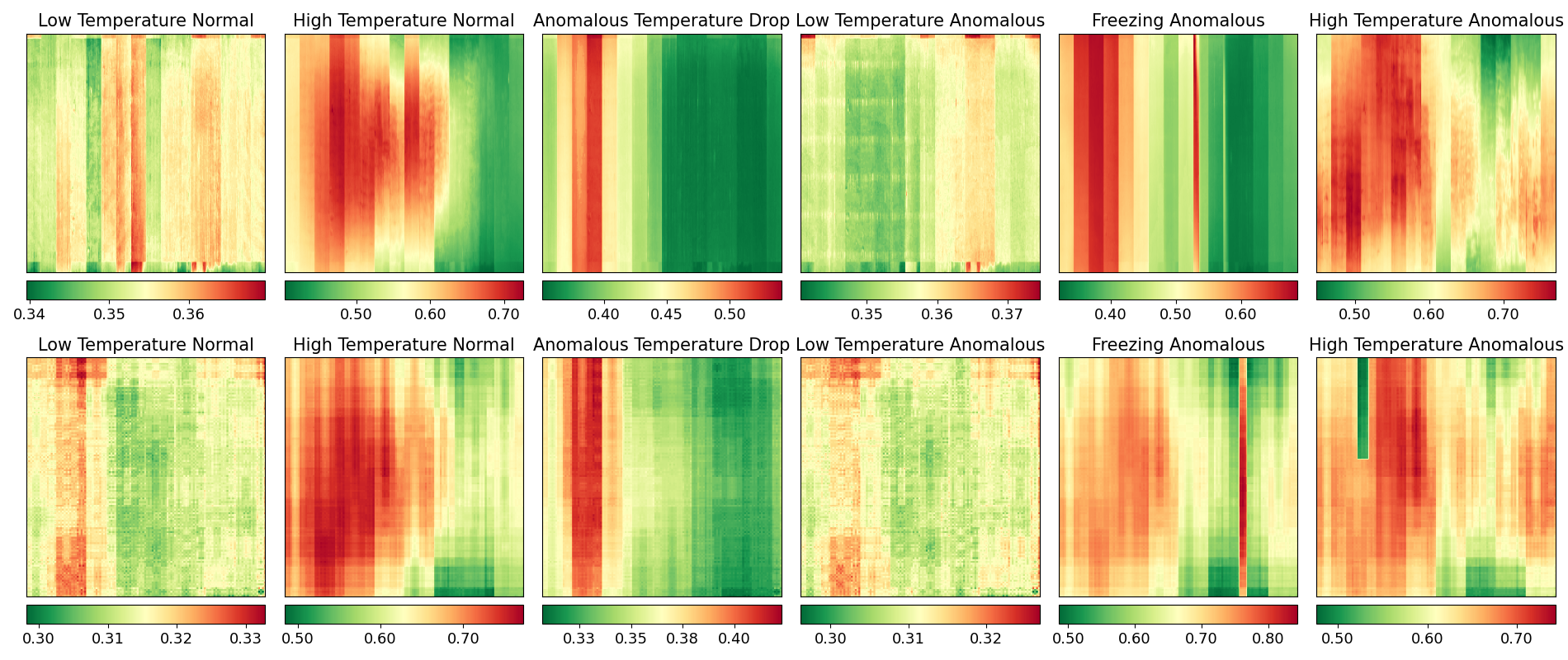}
   \caption{Examples of different types of images in Original (top) and simulated (bottom) dataset}\vspace{1em}
   \label{fig:samples_comparison}   

    \includegraphics[width=\linewidth]{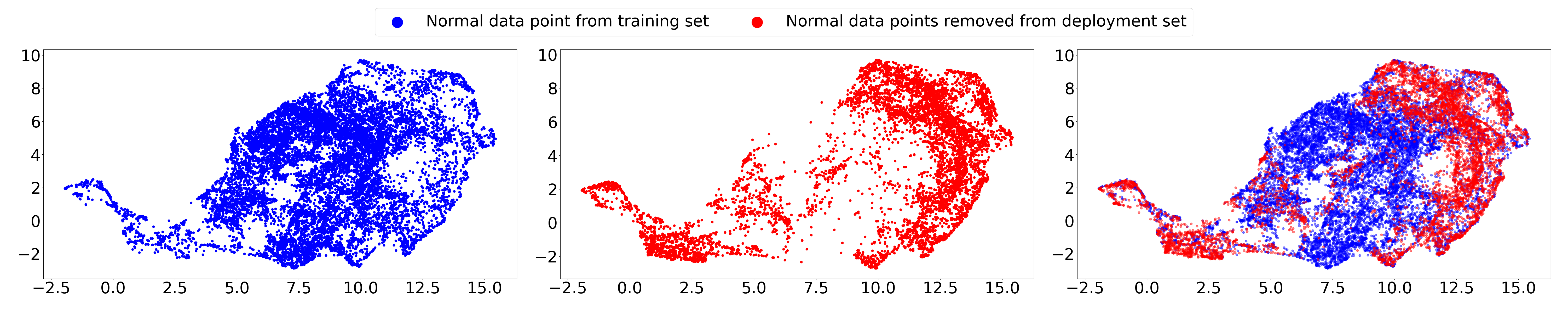}
    \caption{\sukanya{UMAP plot of training set normal images (left), removed deployment normal images (middle) and the both (right).}}
    \label{fig:labelling_cleaning}
\end{figure*}

\sukanya{Some images in the deployment set are wrongly labelled as normal. Such inconsistencies result in a distribution shift between the labelled training and deployment sets, leading to significant degradation of model performance. Thus, we cleaned the deployment set by calculating the distance of each labelled normal image \(x_i\) in the deployment set to the images in the training set \(\mathcal{D}_N\). For this, we first obtain the context embedding \(c_i\) for each image \(x_i\) using the context encoder of \ourmethod{}. We then compute the distance between context embeddings as \(\xi_i = \min_{j = 1, \ldots, |\mathcal{D}_N|} \|c_i - c_j\|_2\). Images from the deployment set for which the distance \(d_i\) exceeds a predetermined threshold are removed. Figure~\ref{fig:labelling_cleaning} shows the UMAP projection \cite{lel2018umap} of the context embeddings corresponding to the samples in the training set \(\mathcal{D}_N\) and the samples from the deployment set that are removed during the cleaning process.}

\section{Additional results}
\label{sec:extended_res}

\begin{table*}[ht]
    \centering
    \caption{Extended Anomaly Detection Performance. Style: \textbf{best} and \underline{second best}}\vspace{-1em}
    \label{tab:res2}
    \aboverulesep = 0pt
    \belowrulesep = 0pt
    \renewcommand{\arraystretch}{1.2}
    \resizebox{\linewidth}{!}{
    \begin{tabular}{@{}c|l|ccc|cccccc|cccccc@{}}
    \toprule
    \multirow{3}{*}{\textbf{Train Setting}} & \multirow{3}{*}{\textbf{Model}} &  \multicolumn{3}{c|}{\textbf{AUROC} (\%)} &  \multicolumn{6}{c|}{\textbf{Accuracy} (\%)} &   \multicolumn{6}{c}{\textbf{F1-score} (\%)}  \\
    &  & \textbf{[Ts\#1]} & \textbf{[Ts\#2]} & \textbf{[Ts\#3]} &  \multicolumn{2}{c}{\textbf{[Ts\#1]}} &  \multicolumn{2}{c}{\textbf{[Ts\#2]}}& \multicolumn{2}{c|}{\textbf{[Ts\#3]}}  &  \multicolumn{2}{c}{\textbf{[Ts\#1]}} &  \multicolumn{2}{c}{\textbf{[Ts\#2]}}& \multicolumn{2}{c}{\textbf{[Ts\#3]}}\\
     &  &  & & & $\lambda_f$ & $\lambda_g$ &  $\lambda_f$ & $\lambda_g$  & $\lambda_f$ & $\lambda_g$ & $\lambda_f$ & $\lambda_g$ & $\lambda_f$ & $\lambda_g$ & $\lambda_f$ & $\lambda_g$\\ 
    \midrule 
    \multirow{15}{*}{\textbf{[Tr\#1]}} 
        &Time of day & 86.99 ($\pm$ 0.00) & 41.44 ($\pm$ 0.00) & 79.97 ($\pm$ 0.00) & 83.03 ($\pm$ 0.00) & 82.86 ($\pm$ 0.00) & 61.60 ($\pm$ 0.00) &  61.60 ($\pm$ 0.00) & 79.55 ($\pm$ 0.00) &   79.41 ($\pm$ 0.00) &   85.34 ($\pm$ 0.00) &      85.17 ($\pm$ 0.00) &     0.00 ($\pm$ 0.00) &        0.00 ($\pm$ 0.00) &     80.20 ($\pm$ 0.00) &        80.02 ($\pm$ 0.00) \\
    &Negative Mean & 81.67 ($\pm$ 0.00) & 46.73 ($\pm$ 0.00) & 71.51 ($\pm$ 0.00) & 70.49 ($\pm$ 0.00) & 76.43 ($\pm$ 0.00) & 10.89 ($\pm$ 0.00) &  13.47 ($\pm$ 0.00) & 60.83 ($\pm$ 0.00) &   66.22 ($\pm$ 0.00) &   74.40 ($\pm$ 0.00) &      76.91 ($\pm$ 0.00) &    14.33 ($\pm$ 0.00) &       12.72 ($\pm$ 0.00) &     65.46 ($\pm$ 0.00) &        66.76 ($\pm$ 0.00) \\
    &Negative STD & 79.31 ($\pm$ 0.00) & 40.76 ($\pm$ 0.00) & 70.35 ($\pm$ 0.00) & 73.60 ($\pm$ 0.00) & 73.60 ($\pm$ 0.00) & 34.10 ($\pm$ 0.00) &  34.10 ($\pm$ 0.00) & 67.19 ($\pm$ 0.00) &   67.19 ($\pm$ 0.00) &   72.49 ($\pm$ 0.00) &      72.49 ($\pm$ 0.00) &    14.18 ($\pm$ 0.00) &       14.18 ($\pm$ 0.00) &     64.66 ($\pm$ 0.00) &        64.66 ($\pm$ 0.00) \\
    &Negative Max & 77.61 ($\pm$ 0.00) & 44.85 ($\pm$ 0.00) & 68.64 ($\pm$ 0.00) & 75.10 ($\pm$ 0.00) & 75.37 ($\pm$ 0.00) & 14.33 ($\pm$ 0.00) &  14.90 ($\pm$ 0.00) & 65.24 ($\pm$ 0.00) &   65.57 ($\pm$ 0.00) &   75.29 ($\pm$ 0.00) &      75.47 ($\pm$ 0.00) &    11.80 ($\pm$ 0.00) &       11.87 ($\pm$ 0.00) &     65.31 ($\pm$ 0.00) &        65.49 ($\pm$ 0.00) \\
    \cdashline{2-17}
    &Autoencoder & 98.05 ($\pm$ 0.74) & 46.43 ($\pm$ 1.61) & 87.87 ($\pm$ 0.26) & 94.50 ($\pm$ 1.22) & 94.49 ($\pm$ 1.22) & 36.96 ($\pm$ 3.72) &  37.19 ($\pm$ 3.68) & 85.17 ($\pm$ 0.46) &   85.20 ($\pm$ 0.47) &   95.27 ($\pm$ 1.06) &      95.26 ($\pm$ 1.06) &    12.58 ($\pm$ 0.64) &       12.62 ($\pm$ 0.63) &     86.45 ($\pm$ 0.54) &        86.46 ($\pm$ 0.55) \\
    &CFlow \cite{Gudovskiy2021CFLOW-AD:Flows} & 94.68 ($\pm$ 1.26) & 39.99 ($\pm$ 2.33) & 82.91 ($\pm$ 1.08) & 87.07 ($\pm$ 1.67) & 86.98 ($\pm$ 1.77) & 31.69 ($\pm$ 1.70) &  32.84 ($\pm$ 1.40) & 78.09 ($\pm$ 1.45) &   78.20 ($\pm$ 1.47) &   88.50 ($\pm$ 1.50) &      88.31 ($\pm$ 1.64) &    12.75 ($\pm$ 0.30) &       12.55 ($\pm$ 0.43) &     79.53 ($\pm$ 1.37) &        79.42 ($\pm$ 1.49) \\
    &Deep SVDD (one-class) \cite{Ruff2018DeepClassification} & 52.76 ($\pm$ 8.71) & 49.22 ($\pm$ 2.66) & 51.85 ($\pm$ 6.15) & 61.35 ($\pm$ 1.95) & 58.34 ($\pm$ 5.62) & 20.00 ($\pm$ 7.87) &  54.10 ($\pm$ 8.04) & 54.65 ($\pm$ 2.86) &   57.65 ($\pm$ 3.52) &   70.79 ($\pm$ 1.77) &      54.97 ($\pm$ 6.57) &    13.79 ($\pm$ 0.87) &       13.33 ($\pm$ 1.53) &     64.12 ($\pm$ 1.60) &        50.54 ($\pm$ 5.49) \\
    &Deep SVDD (soft-boundary) \cite{Ruff2018DeepClassification} & 30.22 ($\pm$ 6.77) & 49.68 ($\pm$ 4.26) & 35.45 ($\pm$ 3.84) & 58.18 ($\pm$ 0.04) & 43.04 ($\pm$ 5.65) &  7.79 ($\pm$ 0.06) &  68.42 ($\pm$ 9.38) & 50.01 ($\pm$ 0.03) &   47.16 ($\pm$ 3.99) &   73.56 ($\pm$ 0.03) &      33.14 ($\pm$ 7.34) &    14.37 ($\pm$ 0.01) &       11.66 ($\pm$ 1.73) &     66.67 ($\pm$ 0.02) &        31.14 ($\pm$ 6.59) \\
    &DR{\AE}M \cite{Zavrtanik2021DRMDetection}& 97.70 ($\pm$ 0.77) & 40.48 ($\pm$ 2.15) & 87.38 ($\pm$ 0.61) & 91.70 ($\pm$ 1.08) & 91.81 ($\pm$ 1.05) & 30.49 ($\pm$ 3.85) &  30.95 ($\pm$ 3.68) & 81.78 ($\pm$ 1.39) &   81.94 ($\pm$ 1.35) &   92.97 ($\pm$ 0.83) &      93.04 ($\pm$ 0.81) &    12.14 ($\pm$ 0.92) &       11.94 ($\pm$ 1.07) &     83.66 ($\pm$ 0.89) &        83.75 ($\pm$ 0.87) \\
    & FastFlow \cite{Yu2021FastFlow:Flows} & 99.83 ($\pm$ 0.03) & 47.32 ($\pm$ 0.29) & 91.36 ($\pm$ 0.25) & 97.39 ($\pm$ 0.29) & 97.38 ($\pm$ 0.29) & 42.12 ($\pm$ 1.27) &  42.18 ($\pm$ 1.29) & 88.43 ($\pm$ 0.38) &   88.43 ($\pm$ 0.38) &   97.72 ($\pm$ 0.26) &      97.71 ($\pm$ 0.26) &    13.22 ($\pm$ 0.26) &       13.24 ($\pm$ 0.25) &     89.17 ($\pm$ 0.35) &        89.17 ($\pm$ 0.35) \\
    &PaDiM \cite{Defard2021PaDiM:Localization} & 99.85 ($\pm$ 0.02) & 49.86 ($\pm$ 0.47) & 91.23 ($\pm$ 0.10) & 96.92 ($\pm$ 0.72) & 96.45 ($\pm$ 0.58) & 43.44 ($\pm$ 1.91) &  44.76 ($\pm$ 1.28) & 88.24 ($\pm$ 0.34) &   88.07 ($\pm$ 0.33) &   97.28 ($\pm$ 0.65) &      96.86 ($\pm$ 0.53) &    13.76 ($\pm$ 0.29) &       13.46 ($\pm$ 0.10) &     88.92 ($\pm$ 0.43) &        88.66 ($\pm$ 0.39) \\
    &PatchCore \cite{Roth2022TowardsDetection} & 99.23 ($\pm$ 0.08) & 50.58 ($\pm$ 0.37) & 89.04 ($\pm$ 0.30) & 95.50 ($\pm$ 0.25) & 95.52 ($\pm$ 0.25) & 31.29 ($\pm$ 1.54) &  31.46 ($\pm$ 1.65) & 85.08 ($\pm$ 0.41) &   85.13 ($\pm$ 0.43) &   96.21 ($\pm$ 0.22) &      96.23 ($\pm$ 0.21) &    15.16 ($\pm$ 0.16) &       15.31 ($\pm$ 0.19) &     86.77 ($\pm$ 0.34) &        86.81 ($\pm$ 0.35) \\
    &Reverse Distillation \cite{deng2022anomaly} & 93.88 ($\pm$ 1.13) & 41.31 ($\pm$ 2.19) & 84.61 ($\pm$ 1.54) & 87.01 ($\pm$ 1.68) & 86.66 ($\pm$ 1.52) & 35.01 ($\pm$ 2.90) &  39.03 ($\pm$ 2.51) & 78.58 ($\pm$ 1.64) &   78.93 ($\pm$ 1.58) &   89.27 ($\pm$ 1.44) &      88.66 ($\pm$ 1.31) &    12.91 ($\pm$ 0.36) &       12.81 ($\pm$ 0.74) &     81.17 ($\pm$ 1.43) &        80.86 ($\pm$ 1.40) \\
    \rowcolor{tabBlue}& \ourmethod{} & 99.86 ($\pm$ 0.05) & 46.22 ($\pm$ 1.06) & 89.89 ($\pm$ 0.35) & 97.73 ($\pm$ 0.34) & 97.65 ($\pm$ 0.27) & 36.10 ($\pm$ 1.19) & 36.62 ($\pm$ 1.35) & 87.73 ($\pm$ 0.29) & 87.75 ($\pm$ 0.26) & 98.04 ($\pm$ 0.29) &    97.97 ($\pm$ 0.24) &  14.02 ($\pm$ 0.50) &     14.13 ($\pm$ 0.50) &   88.76 ($\pm$ 0.25) &      88.75 ($\pm$ 0.22) \\
    \midrule
    \multirow{15}{*}{\textbf{[Tr\#2]}} 
    &Time of day &  86.99 ($\pm$ 0.00) & 41.44 ($\pm$ 0.00) &  79.97 ($\pm$ 0.00) & 83.03 ($\pm$ 0.00) &  82.86 ($\pm$ 0.00) & 61.60 ($\pm$ 0.00) &  61.60 ($\pm$ 0.00) & 79.55 ($\pm$ 0.00) &   79.41 ($\pm$ 0.00) &   85.34 ($\pm$ 0.00) &      85.17 ($\pm$ 0.00) &     0.00 ($\pm$ 0.00) &        0.00 ($\pm$ 0.00) &     80.20 ($\pm$ 0.00) &        80.02 ($\pm$ 0.00) \\
    &Negative Mean &  81.67 ($\pm$ 0.00) & 46.73 ($\pm$ 0.00) &  71.51 ($\pm$ 0.00) & 71.16 ($\pm$ 0.00) &  77.43 ($\pm$ 0.00) &  9.46 ($\pm$ 0.00) &  43.27 ($\pm$ 0.00) & 61.15 ($\pm$ 0.00) &   71.89 ($\pm$ 0.00) &   76.17 ($\pm$ 0.00) &      75.96 ($\pm$ 0.00) &    14.59 ($\pm$ 0.00) &       13.91 ($\pm$ 0.00) &     67.24 ($\pm$ 0.00) &        68.54 ($\pm$ 0.00) \\
     &Negative STD &  79.31 ($\pm$ 0.00) & 40.76 ($\pm$ 0.00) &  70.35 ($\pm$ 0.00) & 66.00 ($\pm$ 0.00) &  73.60 ($\pm$ 0.00) &  9.17 ($\pm$ 0.00) &  45.56 ($\pm$ 0.00) & 56.78 ($\pm$ 0.00) &   69.05 ($\pm$ 0.00) &   74.45 ($\pm$ 0.00) &      70.98 ($\pm$ 0.00) &    12.67 ($\pm$ 0.00) &       12.84 ($\pm$ 0.00) &     66.33 ($\pm$ 0.00) &        64.16 ($\pm$ 0.00) \\
     &Negative Max &  77.61 ($\pm$ 0.00) & 44.85 ($\pm$ 0.00) &  68.64 ($\pm$ 0.00) & 78.37 ($\pm$ 0.00) &  74.71 ($\pm$ 0.00) & 33.81 ($\pm$ 0.00) &  49.86 ($\pm$ 0.00) & 71.14 ($\pm$ 0.00) &   70.68 ($\pm$ 0.00) &   77.19 ($\pm$ 0.00) &      72.26 ($\pm$ 0.00) &    14.13 ($\pm$ 0.00) &       14.63 ($\pm$ 0.00) &     68.62 ($\pm$ 0.00) &        65.87 ($\pm$ 0.00) \\
     \cdashline{2-17}
     &Autoencoder &  96.67 ($\pm$ 0.77) & 45.92 ($\pm$ 2.47) &  85.45 ($\pm$ 1.18) & 93.09 ($\pm$ 1.21) &  93.16 ($\pm$ 1.40) & 26.02 ($\pm$ 3.26) &  28.83 ($\pm$ 2.91) & 82.21 ($\pm$ 1.27) &   82.72 ($\pm$ 1.39) &   94.15 ($\pm$ 1.02) &      94.10 ($\pm$ 1.22) &    13.99 ($\pm$ 0.36) &       13.69 ($\pm$ 0.52) &     84.26 ($\pm$ 1.01) &        84.40 ($\pm$ 1.22) \\
     &CFlow \cite{Gudovskiy2021CFLOW-AD:Flows} &  84.91 ($\pm$ 2.72) & 42.90 ($\pm$ 2.71) &  77.38 ($\pm$ 2.98) & 76.78 ($\pm$ 3.03) &  77.15 ($\pm$ 2.29) & 29.86 ($\pm$ 2.77) &  39.77 ($\pm$ 4.33) & 69.17 ($\pm$ 2.77) &   71.09 ($\pm$ 2.36) &   81.93 ($\pm$ 2.03) &      80.16 ($\pm$ 1.93) &    12.83 ($\pm$ 0.93) &       12.35 ($\pm$ 1.22) &     74.43 ($\pm$ 1.83) &        73.19 ($\pm$ 1.94) \\
     &Deep SVDD (one-class) \cite{Ruff2018DeepClassification} &  45.93 ($\pm$ 5.07) & 52.24 ($\pm$ 2.24) &  46.56 ($\pm$ 4.80) & 58.11 ($\pm$ 0.07) &  54.04 ($\pm$ 4.88) &  7.85 ($\pm$ 0.07) &  61.60 ($\pm$ 2.26) & 49.96 ($\pm$ 0.06) &   55.27 ($\pm$ 4.34) &   73.51 ($\pm$ 0.06) &      47.40 ($\pm$ 4.76) &    14.38 ($\pm$ 0.01) &       12.36 ($\pm$ 1.67) &     66.63 ($\pm$ 0.06) &        44.36 ($\pm$ 4.31) \\
     & Deep SVDD (soft-boundary) \cite{Ruff2018DeepClassification} & 29.21 ($\pm$ 15.05) & 52.48 ($\pm$ 2.83) & 35.44 ($\pm$ 11.72) & 62.34 ($\pm$ 4.10) & 38.41 ($\pm$ 11.14) & 17.65 ($\pm$ 9.91) &  65.04 ($\pm$ 4.93) & 55.09 ($\pm$ 5.05) &   42.73 ($\pm$ 9.28) &   74.76 ($\pm$ 1.16) &      46.42 ($\pm$ 8.16) &    14.06 ($\pm$ 0.30) &       14.09 ($\pm$ 0.89) &     68.12 ($\pm$ 1.42) &        44.14 ($\pm$ 7.40) \\
     &DR{\AE}M \cite{Zavrtanik2021DRMDetection} &  93.52 ($\pm$ 0.52) & 40.51 ($\pm$ 1.33) &  85.71 ($\pm$ 0.78) & 85.00 ($\pm$ 1.41) &  87.14 ($\pm$ 0.98) & 28.54 ($\pm$ 4.62) &  35.70 ($\pm$ 3.60) & 75.85 ($\pm$ 1.66) &   78.80 ($\pm$ 0.98) &   88.07 ($\pm$ 1.00) &      88.98 ($\pm$ 0.88) &    11.84 ($\pm$ 0.36) &       11.40 ($\pm$ 0.36) &     79.62 ($\pm$ 1.04) &        80.67 ($\pm$ 0.85) \\
     &FastFlow \cite{Yu2021FastFlow:Flows} &  92.38 ($\pm$ 0.72) & 52.51 ($\pm$ 1.09) &  89.92 ($\pm$ 0.68) & 88.04 ($\pm$ 0.78) &  87.83 ($\pm$ 0.87) & 56.62 ($\pm$ 3.06) &  59.03 ($\pm$ 2.49) & 82.95 ($\pm$ 1.02) &   83.16 ($\pm$ 1.05) &   90.15 ($\pm$ 0.61) &      89.86 ($\pm$ 0.68) &    13.39 ($\pm$ 0.46) &       13.18 ($\pm$ 0.38) &     84.49 ($\pm$ 0.78) &        84.45 ($\pm$ 0.84) \\
     &PaDiM \cite{Defard2021PaDiM:Localization} &  95.99 ($\pm$ 0.37) & 58.14 ($\pm$ 1.00) &  92.28 ($\pm$ 0.32) & 88.77 ($\pm$ 0.26) &  87.58 ($\pm$ 0.26) & 46.25 ($\pm$ 3.26) &  65.73 ($\pm$ 1.90) & 81.88 ($\pm$ 0.48) &   84.03 ($\pm$ 0.38) &   90.92 ($\pm$ 0.24) &      88.87 ($\pm$ 0.27) &    14.81 ($\pm$ 0.39) &       17.14 ($\pm$ 1.42) &     84.09 ($\pm$ 0.21) &        84.07 ($\pm$ 0.27) \\
     &PatchCore \cite{Roth2022TowardsDetection} &  96.78 ($\pm$ 0.57) & 60.15 ($\pm$ 1.82) &  91.38 ($\pm$ 0.42) & 90.67 ($\pm$ 1.24) &  90.34 ($\pm$ 1.30) & 55.82 ($\pm$ 4.20) &  56.85 ($\pm$ 4.14) & 85.02 ($\pm$ 0.61) &   84.91 ($\pm$ 0.63) &   91.84 ($\pm$ 1.20) &      91.49 ($\pm$ 1.26) &    17.86 ($\pm$ 0.88) &       17.86 ($\pm$ 0.97) &     85.71 ($\pm$ 0.79) &        85.49 ($\pm$ 0.83) \\
     &Reverse Distillation \cite{deng2022anomaly} &  87.19 ($\pm$ 0.99) & 57.22 ($\pm$ 5.77) &  84.04 ($\pm$ 1.64) & 84.39 ($\pm$ 1.15) &  82.48 ($\pm$ 1.14) & 44.41 ($\pm$ 6.44) &  57.02 ($\pm$ 5.75) & 77.91 ($\pm$ 1.16) &   78.36 ($\pm$ 0.97) &   87.32 ($\pm$ 0.91) &      84.92 ($\pm$ 1.12) &    14.91 ($\pm$ 1.50) &       15.50 ($\pm$ 1.41) &     80.60 ($\pm$ 0.71) &        79.53 ($\pm$ 0.91) \\
   \rowcolor{tabBlue} & \ourmethod{} & 94.78 ($\pm$ 1.09) & 85.81 ($\pm$ 1.23) & 92.53 ($\pm$ 0.81) & 88.82 ($\pm$ 1.33) & 88.75 ($\pm$ 1.28) &  76.68 ($\pm$ 2.13) &  78.40 ($\pm$ 1.36) & 86.85 ($\pm$ 0.89) & 87.07 ($\pm$ 0.98) & 90.03 ($\pm$ 1.24) &    89.88 ($\pm$ 1.19) &  35.40 ($\pm$ 1.59) &     36.29 ($\pm$ 1.40) &   86.83 ($\pm$ 1.01) &      86.89 ($\pm$ 1.06) \\

    \bottomrule
   \end{tabular}
   }
\end{table*}

\textbf{Simple baselines}. \sukanya{We also consider four simple baselines which compute thresholds based on the features extracted from the samples in the validation set to detect anomalies. The extracted features can be seen as the anomaly scores. In this study, we consider the following features:
\begin{itemize}
    \item Time of day: the time when the image is recorded.
    \item Negative mean: negative of the average pixel value of the image.
    \item Negative max: negative of the maximum pixel value of the image.
    \item Negative Std: negative of the standard deviation of the pixel value in the image. 
\end{itemize}}

\textbf{Threshold selection}. \sukanya{Given the anomaly scores assigned to the labelled validation samples, we explore two different threshold selection approaches:
\begin{itemize}
    \item \textbf{F1-score}. We compute F1 scores by considering various thresholds. Given the F1 scores, we select the optimal threshold $\lambda_f$ as the one that corresponds to the maximum F1 score.
    \item \textbf{G-Mean}. Similar to $\lambda_f$, we calculate the G-Mean value by applying multiple thresholds The optimal threshold $\lambda_g$ corresponds to the threshold where we obtain the highest G-Mean value. G-Mean is the geometric mean of specificity and recall.
    \begin{equation*}
        \text{G-Mean} = \sqrt{\text{Specificity} \cdot \text{Recall}} = \sqrt{(1-FPR) \cdot TPR}
    \end{equation*}
\end{itemize}}

Based on the selected threshold, Accuracy and F1-score are reported for \textbf{[Tr\#1]} and \textbf{[Tr\#2]} over three test setups in Table \ref{tab:res2}.

\end{document}